\documentclass{article}
\usepackage{siunitx}
\usepackage{csquotes}
\usepackage[inline]{enumitem}
\usepackage{graphicx}
\usepackage{xspace}
\usepackage{amssymb}
\usepackage{amsmath}
\usepackage{ntheorem}
\usepackage{stmaryrd}
\usepackage{hyperref}

\input{sty/shortcuts.sty}
\input{sty/trcommands.sty}
\title{Framing Relevance for Safety-Critical Autonomous Systems}
\author{Astrid Rakow\thanks{This research was supported by the German Research Council (DFG) in the PIRE Projects SD-SSCPS and ISCE-ACPS under grant no. DA 206/11-1.}\\a.rakow@uol.de\\Carl von Ossietzky University of Oldenburg}
\begin{document}
\maketitle
\begin{abstract}We are in the process of building complex highly autonomous systems that have build-in beliefs, perceive their environment and exchange information. 
These systems construct their respective world view and based on it they plan their future manoeuvres, i.e., they choose their actions in order to establish their goals based on their prediction of the possible futures. 
Usually these systems face an overwhelming flood of information provided by a variety of sources where by far not everything is relevant.
The goal of our work is to develop a formal approach to determine what is relevant for a safety critical autonomous system (\HAS) at its current mission, i.e., what information suffices to build an appropriate world view to accomplish its mission goals.
\end{abstract}
\tableofcontents
\section{Introduction}
Full informedness is certainly not necessary for successful manoeuvres of highly autonomous systems.  
For instance, when an autonomous car approaches a pedestrian crossing, it has to decelerate if  pedestrians want to cross the road --irrespectively of their shirt colours or the exact number of pedestrians. Nevertheless, the number of pedestrians is relevant for the expectation when the group will have crossed the road, influencing the decision whether to take a detour circumventing the crossing. 
For the latter case, the relevance of the group size results from the goal of minimising the travel time.

\begin{figure}
	\includegraphics[width=\textwidth]{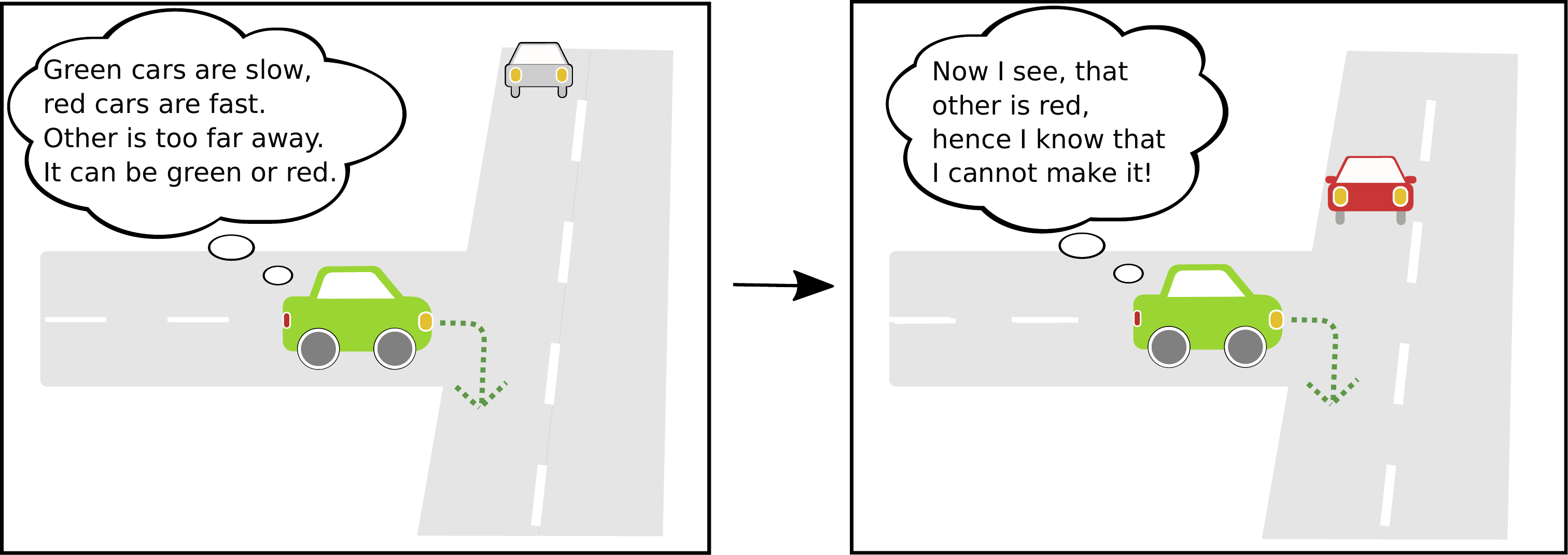}
	\caption{The autonomous system \ego wants to save time, but, even more, it wants to avoid collisions. It has to decide whether to do the right turn before the car \other has passed. \Ego's color perception of distant objects is not working.}\label{fig:comic}
\end{figure}

The control of an autonomous system \HAS can be considered as implementation of a strategy that chooses control actions (time bounded services provided by its autonomous layer like "follow the lane and accelerate" or "emergency braking")  based on the currently agglomerated information. 
A decision for an action is based on the combination of \HAS's observations of the world and \HAS's insights into the world -- e.g. \HAS observed the upper speed limit and knows about the effect of acceleration on its speed. 
Since \HAS usually has only limited sensing and communication capabilities and hence limited means to assess the situation, it faces uncertainties in determining its current situation. Several alternative worlds seem possible at a time and it cannot tell which one actually represents the reality best (cf. \autoref{fig:comic}).

Our research is driven by the question "What does an autonomous system \HAS need to perceive and know for a successful autonomous manoeuvre, i.e. for manoeuvres where it takes decisions based on its beliefs?". 
We thereby strive to define a notion of relevance of observations and knowledge \footnote{We should rather write \enquote{hard believes}, since the autonomous system has no mean to access the ground truth and hence only treats certain propositions as knowledge.} for autonomous strategic decisions. 
To pave the way for a formal treatment of this question, we develop in \cite{doxFrame} a formal model that explicitly represents the beliefs of \HAS. 
Within this framework, we characterise \emph{autonomous-decisive systems} as systems that rationally take decisions based on the content of beliefs.
We regard a system as rational, when it chooses actions that it believes promise success. 
In this report, we define \emph{relevance} of observations and knowledge for an autonomous-decisive system \HAS with its goals \goalList. 
Basically, a combination of knowledge, observations and possible beliefs is relevant if we cannot omit anything of them while being equally successful.
We present an algorithmical approach based on strategy synthesis to determine relevant combinations of knowledge, observations and possible beliefs. 
Conceptually, the presented approach will be useful in the early design of \sys, where simple and abstract models are considered.
We assume that a design-time world model \universeD is given that characterises the application domain and captures test criteria of \HAS. 
Such a world model may be derived from scenario-databases and test catalogues. 
We moreover assume that prior to our outlined analysis, the scope of beliefs (= set of the possible beliefs)  has been defined. So it has been defined which artefacts, objects, and interrelations will possibly be represented in \HAS's beliefs. 
We envision that the starting point for this design step could be \cite{DF11}.

We expect that this work may help to guide the design of beliefs and highlight the trade-off between sensing capabilities (including communications) and  knowledge about the world.

Later design steps  will have to generalise \sys's capabilities to deal with the known unknown aspects of design-time world, taking into account that no world model will match the reality. 
The sanity of derived beliefs will also be judged regarding its robustness against the unknown.
We consider these aspects as future work.

Our notion of relevance \enquote{Relevant is what is necessary to know or to perceive in order to perform best} is based on the dynamically formed beliefs of the system \HAS and is thus subjective, dynamic, motivational and cognitive.
To formalise autonomous decision making and our notion of relevance we use a doxastic model, i.e. a model that explicitly captures beliefs using possible world semantics. 
While many approaches in literature, e.g. \cite{EpistemicsLogics,DoxasticMeyer2003}, regard a possible world as a single \enquote{flat} node, here a possible world has an inner structure. A possible world is a Kripke structure itself, that captures the past, presence and extrapolated future as imagined by \HAS and thereby explains its autonomous decisions.

In \cite{DF11} Damm and Finkbeiner determine the optimal perimeter of a world model as the subset of a Kripke structure's propositions that is necessary to synthesize a winning strategy. 
We generalize their idea in order to define relevance for \HASes. 
To this end, we distinguish between the model of ground truth design time model and the model of beliefs, based on which \HAS take decision which in turn effect the ground truth. 

\paragraph{Outline}
In the next section we discuss related work concerning the notion of relevance. The notion of relevance has been discussed in many fields of science, but probably most prominently in information retrieval and information science. Although IR and autonomous system design might seem very different in nature, much of the foundational work in IR regarding the notion relevance finds application also for determining what is relevant for a autonomous system.
We  present the framework within which we capture our notion of relevance  
in \autoref{sec:relFrameIR} and \autoref{sec:ingredients}. 
The latter section and \autoref{sec:autonom} on the notions of autonomous and automatic systems follow closely \cite{doxFrame} which is previous work published under the Creative Commons Attribution License. 
In this paper additional material can be found as well as sleeker proofs. \autoref{sec:autoSys} is a new addition to \autoref{sec:autonom}.  
In \autoref{sec:relevance} we develop our notion of relevance for safety-critical autonomous decisions.

\section{From Relevance in IR to Relevance for Autonomous Safety-Critical Systems}\label{sec:relevanceIR}\label{sec:relatedWork}
\subsection{Relevance in IR}
Although relevance is discussed in  many fields such as
philosophy, psychology or artificial intelligence, it is probably most prominently discussed in information science and information retrieval (IR) 
where it is considered to be among the most central challenges  \cite{historyRelevance,relevanceFramework,RelevanceRelation,ReExORel}. 
Our notion of \enquote{relevance of perceptions and knowledge of an autonomous safety-critical system \sys} is related in many ways to the notion of relevance in IR.
The later notion has its beginnings in times when librarians without computer support were trying to retrieve documents for their customers ~\cite{historyRelevance}. 
Although the task of retrieving relevant documents may seem quite different from determining what input an autonomous system needs in order to be successful, an abstract concept of relevance should be applicable to both fields alike.
Hence especially the foundational work on relevance in IR remains valid or analogies can be drawn for relevance for autonomous safety-critical systems. 
Even the more so, when we consider relevance as discussed in \cite{MobiUbi,Reichenbacher2007,DeSabbata2015} in the rather young field of mobile IR systems. 

We will discuss the relation between relevance in IR and relevance for an autonomous safety-critical system \HAS later in \autoref{sec:relHAS}. 
Here we first give a short introduction to the concept of relevance in IR and then present a condensed overview of its history.
Since the literature on relevance is vast, we do not claim to give a complete overview. The following is meant to give an introduction to relevance in IR with a focus on the line of research closest to ours.
\subsubsection{What is Relevance in IR?}\label{sec:relIR}
We feel urged to remark that there is not \emph{the} notion of relevance in IR,  but there is an agreement that relevance is a relationship -- basically between a document and
am information need \cite{historyRelevance}. 
The quest of understanding the nature of relevance lead to various definitions since the 1960s.
One reason for this still ongoing quest is that \enquote{relevance is not a single
notion, but many} as Wilson stated in 1973~\cite[p.457]{Wilson1973}. 
Saracevic  remarks in his influential survey \cite{relevanceFramework} that \enquote{In the most fundamental sense, relevance has to do with effectiveness of communication}~\cite[p.\ 321]{relevanceFramework}. 
He developed in \cite{Saracevic1996} a stratified system of relevance distinguishing \emph{system relevance}, \emph{topical relevance}, \emph{cognitive relevance}, \emph{situational relevance} and \emph{motivational relevance} (cf.~\autoref{tab:stratified}).
According to Saracevic the different strata dynamically interact and are interdependent.  
\emph{Topicality}, the quality of a document to convey information about the topic of the information need, lies at the heart of relevance \cite{historyRelevance,RelevanceRelation}. 
\begin{table}[htbp]
	\footnotesize
	\begin{tabular}{l p{10cm}}
		Relevance: &  Relation between \ldots\\
		\hline\\
		System  & a query and information objects (texts) in a
		collection as retrieved or as failed to be retrieved.\\
		Topical &  the topic expressed in a query and the topic covered by the retrieved texts.\\
		Cognitive & the state of knowledge and cognitive information need of a user and the retrieved texts.\\
		Situational & the situation, task, or problem and the retrieved texts.\\
		Motivational & the intent, goal, and motivation of a user and the retrieved texts.\\
	\end{tabular}
	\caption{Stratified system of relevance by Saracevic \cite{Saracevic1996} according to \cite{RelevanceRelation} }\label{tab:stratified}
	\normalsize
\end{table}

\begin{figure}[h]
	\centering
	\includegraphics[width=0.9\textwidth]{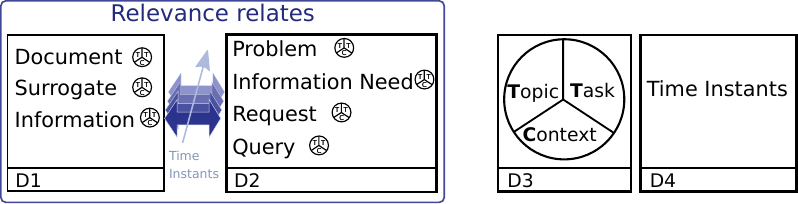}
	\caption{Mizzaro's four-dimensional framework for relevance in IR\cite{historyRelevance}; \footnotesize D1:  the (physical) document $d$; the surrogate $s$, which is a representation of $d$; the information $i$, which the user receives via reading $d$. D2: the problem $P$ the user is facing; the information need $N$ that is represented in the user's mind; the request $R$, which is a representation of $N$ in  a human language; the query $Q$, which is a representation of $R$, in a system language. The  entities $d,s,i$ of D1 and $P,N,R,Q$ of D2 can be decomposed into (D3a: the topic, that is the subject area, the user is interested in), (D3b: the task, that is the activity that the user will execute with the retrieved information) and (D3c: the context, which includes everything not pertaining to topic and task).} \label{fig:relevanceRelation}
\end{figure}

Mizzaro presented in \cite{historyRelevance} his four dimensional model of relevance recognizing that relevance also has a time dimension. Relevance is still regarded as a relation between two entities of the two groups D1 (document/surrogate(information)  and D2 (problem/information need/request/query) (cf. \autoref{fig:relevanceRelation}). As third dimension he considers D3 (topic/task/context). But since the user perceives the problem in a different way over time, the fourth dimension of Mizzaro's framework are \enquote{the various time instants from the arising problem until its solution}~\cite[p. 812]{historyRelevance}.

In \cite{DeSabbata2015} De Sabbata et al. apdapt Mizzaro's framework to describe relevance for mobile IR systems. 
They observe that for such systems where the user might be moving (i) \emph{the relation between space and time} and (ii) \emph{a link to the real world} is an important factor for the relevance of the retrieved information. 
The user's information need might originate at a location $l$, there information $i_l$ would be relevant, but $i_l$ may have ceased to be relevant to user, when retrieved, since the user is then at another location $l'$ \cite{Reichenbacher2011,DeSabbata2015}. 
To emphasize the spatio-temporal nature of the information seeking they introduce the new 'space-time dimension'. 
Furthermore, they introduce 'world' as a another new dimension, in order to capture the influence of different abstractions of reality.
They argue that since the real information need is different from the query received by the system and the real world is different
from the world perceived by the system, a relevance concept can be described as dealing with reality at the different abstraction levels: the real world; the documented world (recorded by the human and stored); the perceived world (perceived by the user); the system world (world as it is known by the system).

To summarize, early on it was recognized that relevance is determined by many factors, which was coined \enquote{multi-dimensionality} of relevance.
At the heart of relevance lies topicality. 
The retrieval process can be considered from the system and the user perspective. 
The situation the user is in, his cognitive state and the his goals and intentions influence what is relevant for him (cf. \autoref{tab:stratified}). 
With the rise applications that retrieve information about the user's surrounding, the link to the real world in space and time gained importance.

%
\subsubsection{A Short History of Relevance in IR}\label{sec:history}
%
This section gives an overview of the history of relevance condensed to the works that we consider especially important with regard to our notion of relevance for an autonomous safety-critical system \HAS. 
We thereby like to stress that early on approaches were developed to formally describe relevance, that more recently there is intensified research on cognitive aspects of relevance, and that due to mobile IR, a strong link to the real world in space and time has been recognized.    
The following short compilation is based on \cite{historyRelevance,RelevanceRelation} and extended by an update.

In the \emph{period 1959–1976} efforts were focused on understanding the nature and conceptual subtleties of relevance and devising definitions using various mathematical tools. 
The main contributions to foundational work were 
\begin{itemize}
	\item \cite{MaronKuhns}: In the year 1960, Maron and Kuhns propose weighted indexing. 
		The computed weights are meant to reflect the probably of the document beeing relevant to that user, so that the documents can be ranked in descending order of predicted relevance. 
	\item \cite{Rees}: 1966, Rees notes that the definition of relevance should reflect the influence of \enquote{the previous knowledge} of the user and the \enquote{usefulness} of the information.
		Relevance is thereby a user construct and highly subjective.

	\item \cite{Cooper}, \cite{Wilson1973}: In 1971, Cooper uses in \cite{Cooper} mathematical logic to define relevance. 
		In particular, Cooper defines that a sentence $s$ is relevant to a sentence $r$ if $s$ belongs to a minimal set of premises $M$ entailing $r$, i.e., $\textit{relevant}(s, r)$ iff $\exists M$ $(s \in M \land M \models r \land M - s \not\models r)$. 
		A document $D = \{s_1 , s_2 , \ldots , s_n\}$ is relevant to a request $r$, $Relevant(D, r)$, iff $\exists i (relevant(s_i ,r))$.
		Thereby Cooper gives rise to the today's notion of \emph{logical relevance}.

		In \cite{Wilson1973} Wilson (1973) tries to improve Cooper's definition by taking into account a user's \enquote{situation}, \enquote{stock of information} and \enquote{goals}.
		Today this kind of relevance is referred to as \emph{situational relevance}%
\end{itemize}
This period ends with the surveys \cite{relevanceFramework,TenYearsSaracevic,Saracevic1970,relevanceFramework} of Saracevic where he summarizes and classifies previous work, laying the bases for future research. 

In the following period, a new stream of works is concerned with the importance of the user.
By means of empirical end-users studies further relevance criteria, apart from topicality, are identified, based on which users judge the relevance of the retrieved such as e.g. recency, quality or verification\cite{RelevanceRelation}. 
During this period, \emph{cognitive relevance} and \emph{situational relevance} are elaborated by e.g.\ \cite{Ingwersen1996,Cosijn2000,Saracevic1996,Hjorland2002,Saracevic2007a,ResonanceRuthven2021}.

A second stream of works, concerned with defining a logic for IR, is triggered by the works of van Rijsbergen \cite{Rijsbergen1986,Rijsbergen89}.
As an example we want to mention in particular \cite{Nie1989}, where Nie formalises relevance via modal logic and
Kripke’s possible world semantics. 
The query is represented by a formula and the document by possible worlds.

A third stream deals with the challenges induced by mobile scenarios and digitalisation.
For this domain the  need of representing the world surrounding a user was recognized~\cite{Mountain2001,Reichenbacher2004,MobiUbi,Mountain2007,Raper2007,Reichenbacher2007,DeSabbata2015}.

Research on the notion of relevance is still ongoing.
Among the open questions is, how are the different dimensions of relevance related? 
Huang \& Soergel remark in \cite{RelevanceRelation} that \enquote{Relevance is still by and large a black box [\ldots] We may be capable of telling whether A is relevant to B, but specifying precisely in what way A is relevant to B is much harder} \cite[p.~32]{RelevanceRelation}
%
%
\subsection{A Notion of Relevance for Autonomous Safety-critical Systems}\label{sec:relHAS}
\begin{figure}	
	\centering
	\includegraphics[width=0.8\textwidth]{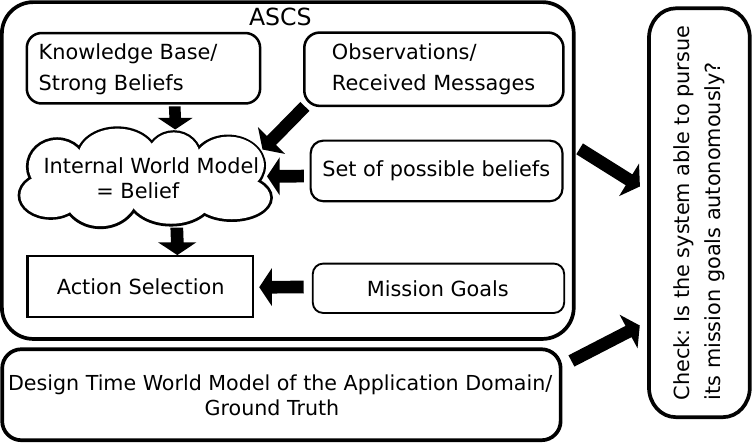}
	\caption{Ingredients of the relevance framework for autonomous safety-critical systems (ASCS)}\label{fig:frame}
\end{figure}

%
%
We now turn to the challenges of defining a notion of relevance for a safety-critical autonomous system \HAS. 
We survey the main differences of \enquote{relevance of perceptions and knowledge and possible beliefs for {\sys}} to the traditional information seeking problem in IR and motivate our conceptual approach to defining relevance. 
Figure \ref{fig:frame} gives an overview of the main ingredients of this approach, as will be discussed in this and the following section.

The notion of relevance in IR originated from the document retrieval problem in libraries. 
In Mizzaro's terms (cf.~\autoref{fig:relevanceRelation}) this process can be described as follows: 
A user has a problem to solve and recognizes an information need. He hence decides to go to a library. There he requests documents. This request can in turn be translated to a system query. The system retrieves the relevant documents for the user, who's information need changes by the retrieved.

In contrast, we are interested in autonomous systems and aim to support the design process of such systems.
We hence do not have a user who formulates a request and there is no retrieval system operating on a database. 
Instead, we assume that the design domain of the system \HAS is known and moreover a list of requirements for \HAS has been defined.
These requirements allow to define missions for \HAS.
For instance an autonomous vehicle can have a mission like
\enquote{drive on a highway from location $l_1$ to $l_2$, master this mission within time $t$, do not exceed the speed limit $v_{max}$, respect the safety distance at all times, by all means avoid severe collisions}. 
So a mission restricts the application domain to a more specific setting and assigns a \emph{prioritized list of mission goals}. 
The overall behaviour of \HAS can be described as a compilation of missions instantiated to the concrete goals and circumstances. 

In contrast to IR systems, where a rather explicit information need has to be satisfied, for an autonomous safety-critical system \HAS  the information need arises from mission goals.
\HAS has to accomplish its mission goals within the real world. 
It therefore \emph{chooses its actions} based on its assessment of the situation, which includes its prediction of the possible future evolutions.
In order to accomplish its goals, \HAS must sufficiently well predict how its actions effect the real world.
Therefore \HAS is equipped with sensors providing perceptions of its environment. 
These are then integrated by \HAS into its \emph{internal world}. 
Similarly, \HAS may receive messages from other agents conveying information. 
In due course, we do not distinguish between perceptions and messages.
So, while in IR documents are retrieved in order to satisfy an explicit information need, for \HAS the information need is implied by its goals and relates to perceptions of the world. 

A user of an IR system is assumed to have stock of prior knowledge and this knowledge evolves during her search. 
We likewise assume that \HAS gets equipped with a so-called knowledge base \kbase during design and that this knowledge base evolves. 
A \HAS may \enquote{forget} certain statements of \kbase and it may gain new statements, that are provided by trusted sources during its mission. 
We assume that the entries of \kbase are \emph{believed knowledge}/\emph{strong beliefs} of \HAS, i.e. it  believes that they are true, but they are not necessarily true. 
The \HAS uses its knowledge base when maintaining its internal world model.
\kbase may hold rules how to combine and integrate perceptions and messages, like \enquote{\emph{Cars drive on the road not under.}}, \enquote{\emph{There is a traffic jam ahead.}}. 
So, for \HAS also relates not only to perceptions of the world but also to believed knowledge. 

Note that the internal world model of \HAS  is its \emph{belief}.  
Hence \HAS chooses its actions based on this belief.

The information need of \HAS can be highly dynamic. 
For instance, an \AV driving along a road has to know about obstacles appearing on its way and about the road conditions at the time. 
Even if the information need has not been satisfied,  \HAS often is forced to choose an action anyway. 
Even if the \AV does not get the information whether the road ahead is slippery, it has to continue to drive, since it cannot and should not stop instantaneously.
So, the systems often choose actions despite uncertainties. 

If an autonomous vehicle rather unexpectedly slips and leaves the road, a sudden reassessment of the situation takes place in order to devise a plan how to ensure its most pressing goals. A replanning has to take place. 
So the prioritized goals imply a prioritized information need.

Similar to mobile IR systems, safety-critical autonomous systems have a strong link to the real world and often the time-space dimension is also very important.
In comparison to mobile IR systems, the dynamicity can be very high for safety-critical autonomous systems. Such a system \sys must be able to suddenly reassess the situation, change its goals and hence its information need and devise a new plan.

Although a safety-critical autonomous \HAS may continuously interact with the real world under high demands an reactivity, usually \HAS does not continuously need to update every aspect of its world model. 
If the road is now slippery and it is a cold and wet January morning, it is sensible to assume that the road still will be slippery in $1ms$.
An engineer might establish the rule as part of  \HAS's knowledge base. This rule constrains what \HAS imagines is possible --\HAS believes only in worlds where the road is now and in the near future slippery.
Additionally, an engineer might decide that misclassifying a giant flower pot as litter bin is tolerable.

\begin{figure}[htbp]
	\centering
	\includegraphics[width=0.5\textwidth]{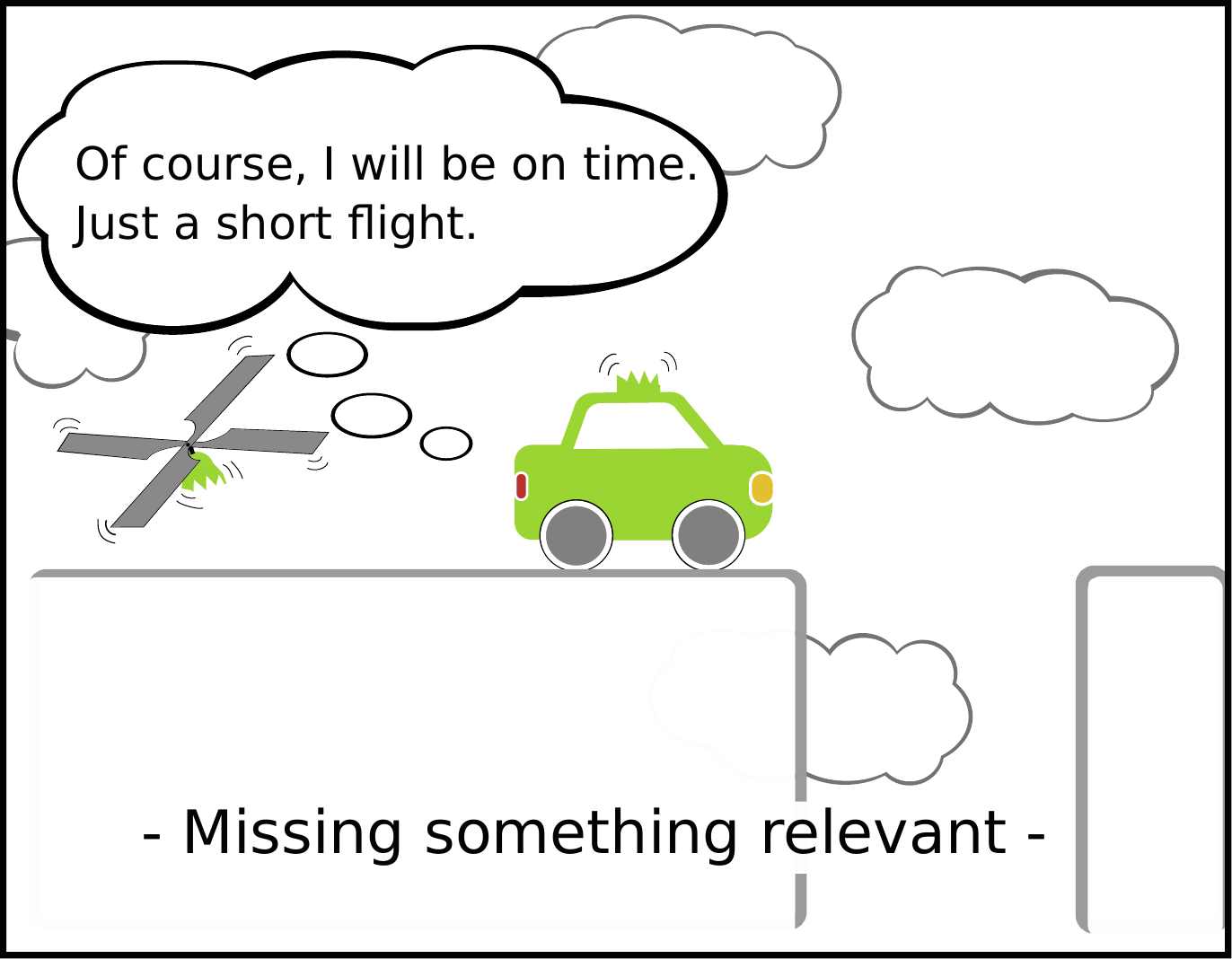}\hspace*{2cm}
	\caption{Relevant observations and knowledge are mission critical. The basic idea for formally defining relevance is that a mission goal cannot be achieved if relevant observations or knowledge is missing.\normalsize
	}\label{fig:fly}
\end{figure}

We conclude that \HAS implements its kind of cognitive process.
This process defines how  built-in knowledge and gained perceptions result in a belief.
Based on its beliefs \HAS decides on its actions, i.e. \HAS \emph{decides autonomously} (a notion that we will formally introduce in \autoref{def:autonomDec} on page \pageref{def:autonomDec}). 
If its beliefs deviate from the real world so that \HAS is not able to achieve its goals in the real world, then \HAS misses some relevant information (cf. \autoref{fig:fly}).
In the following we will develop a formal framework based on this concept.

%
\subsection{Our Relevance Framework within the Design Process}\label{sec:designproc}
%
Goal of this line of work is to develop  a formal approach to determine what knowledge and observations are relevant for a safety-critical autonomous \HAS at its mission.
Thereby we aim to support an engineer that has to decide at design-time what sensors and processing power \HAS gets and how it constructs its beliefs, such that \HAS will be able to accomplish its mission goals.

We assume that the engineers capture the application domain (including test criteria of \HAS) via a formal model of the world at design time. 
We refer to this design-time model as $\universeD$. 
We will use $\universeD$ within our framework as an anchor to judge what is relevant for the real world. We refer to it also as \emph{ground truth} (cf. \autoref{fig:frame}).

\begin{figure}[htbp]
	\centering
	\includegraphics[width=\textwidth]{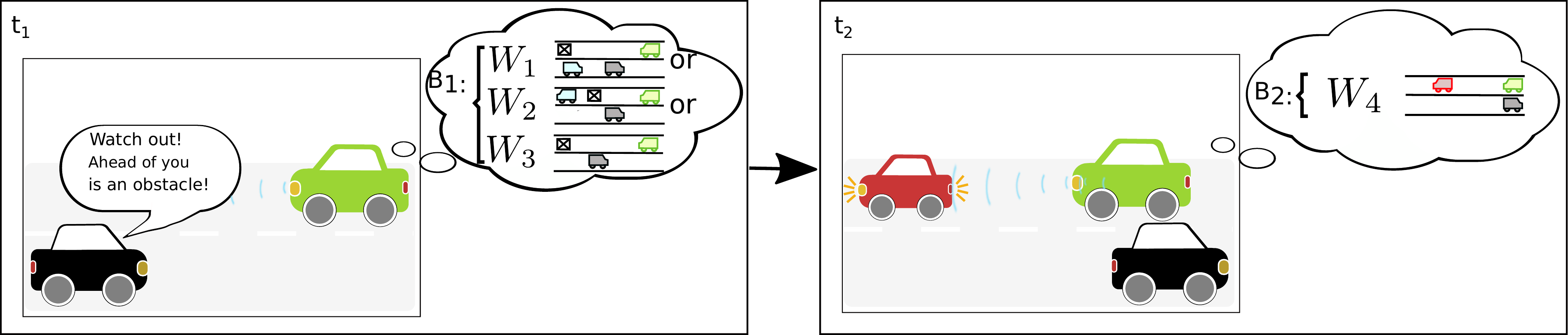}\hspace*{2cm}
	\caption{Possible World Semantics. \footnotesize The green car believes at time $t_1$ that there is an object somewhere and hence thinks that $W_1$ or $W_2$ or $W_3$ are possible worlds. With other words, its belief is $B_1$ according to which $W_1$ in state $\state_1$ or $W_2$ in state $\state_2$ or $W_3$ in state $\state_3$ are possible. At time $t_2$ it has perceived the broken car ahead and updates its belief to $B_2$ containing only the possible world $W_4$ in state $\state_4$. normalsize }\label{fig:possibleWorldsSemantics}
\end{figure}

We assume that the engineer has determined which artifacts the system \HAS must represent in order to build up an internal representation of the real world, a world model \world.
Since the resources of \HAS are finite, we consequently assume that there are only finitely many different world models \HAS can possibly represent.
The set of all possible world models is denoted as \Worlds.
At a time instance \HAS may deem several world models possible and it imagines itself to be currently at certain states of these worlds. 
Thus, to describe a \emph{belief} of \HAS  we use the possible world semantics. 
A world describes not only the current situation, that is the current state. 
It describes the involved objects and what they can do, how they interact, where they come from and what might happen in the future. 
\autoref{fig:possibleWorldsSemantics} illustrates the terms on a abstract example. We will formally introduce them in \autoref{sec:relFrameIR}.

As motivated in the previous section, we moreover assume that a system \HAS has a \emph{knowledgebase} \kbase representing the knowledge built in during the design. 

We have now informally introduced the ingredients of our framework as depicted in \autoref{fig:frame}. 
Based on the built-in knowledge and its perceptions \Obs of the real world, \HAS constructs and maintains its beliefs and autonomously decides based on its beliefs, what to do. 
Within this framework we formalize, that relevant is what \HAS needs to know and observe
to form believes that enable it to act successfully in $W_d$.
Formally, we represent the construction/maintenance of beliefs via a belief formation (function). 
The belief formation \LabelB defines the current belief of \HAS. 

\emph{In a nutshell, our approach simulates what \HAS thinks when performing its maneuver in \universeD and what it does due to its beliefs.  
The criteria for having the relevant observations and the relevant hard beliefs and sufficient possible beliefs is whether \HAS achieves its goals -- or more precisely whether \HAS is able to form beliefs based on the observations and knowledge based on which it can achieve its goals.}

\begin{figure}
	\centering
	\includegraphics[width=0.5\textwidth]{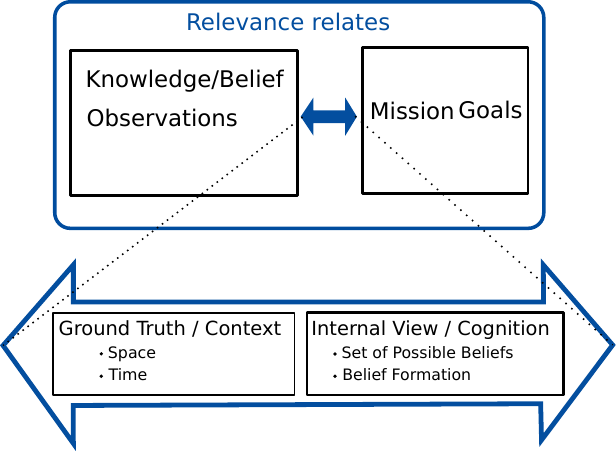}
	\caption{A framework for relevance for autonomous safety-critical systems ASCS}\label{fig:relevanceRelationHAS}
\end{figure}

\begin{figure}[htbp]
	\centering
	\includegraphics[width=0.8\textwidth]{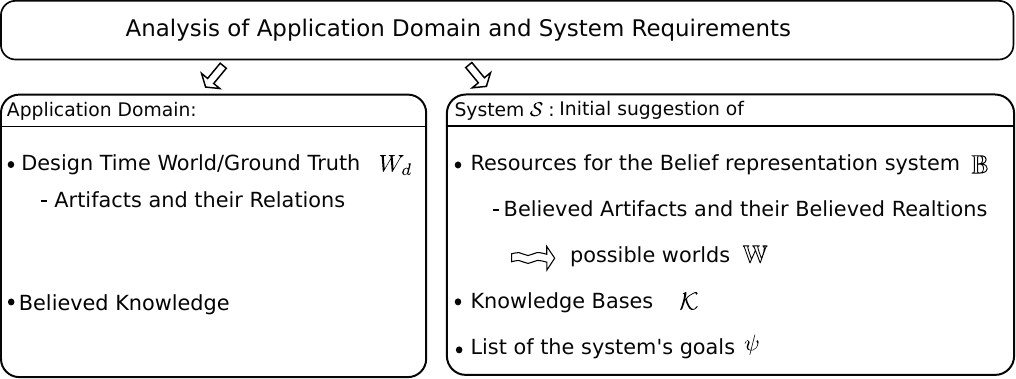}
	\caption{Input for the relevance framework from design process and the resulting formal ingredients}\label{fig:design}
\end{figure}

In contrast to Mizzaro's framework (cf.~\autoref{fig:relevanceRelation}) our notion of relevance has a strong emphasis on the cognitive dimension of relevance, since we treat belief formation as a central ingredient of our framework. \autoref{fig:relevanceRelationHAS} illustrates this conceptual difference. 
\autoref{fig:design} illustrates that our approach aims to support the early design. We assume that an analysis of the application domain and \HAS's requirements has been done.

Apart from providing a characterisation of what knowledge and perceptions are relevant, the framework contributes to tackling the following questions:
\begin{enumerate}
	\item\label{i:auto} Given ground truth, the set of possible beliefs, observations, knowledge, goals,
		is it possible that \HAS forms a belief based on which it performs successful?
	\item\label{i:jitter} Given ground truth, the set of possible belief universe, observations, knowledge, goals, belief formation,
		will \HAS perform successfully?
		How much jitter of perceived values is tolerable, how relevant is the exact timing, the assumed dynamics?
	\item\label{i:sharing} Given ground truth, the set of possible beliefs, observations, knowledge, goals, belief formation, a partition of percepts and required time separation of these partitions,
		will \HAS perform successfully without relying on perceptions violating the required time separation? 
\end{enumerate}
Question \ref{i:auto} occurs during the design, after the domain and requirements analysis (cf. \autoref{fig:design}). Given the application domain has been analysed and a formal model of it exists, the sensory input of \HAS has abstractly been defined, an initial proposal for the build-in knowledge and for the inner representation of the world of \HAS has been made. Then we can examine whether \HAS can somehow build and maintain a belief, that is sufficient to act successfully within the assumed world. Since beliefs are a coarse approximation --e.g. due to limited storage and computation resources-- and the belief formation may be in parts \enquote{wrong} --e.g. since it is not possible to observe certain aspects of the world--, this is an interesting question.   

Later in the design, after fixing the way \HAS constructs its inner world representation, question \ref{i:jitter} arises. 
In contrast to question \ref{i:auto}, it evaluates whether a given belief formation is sufficient for \HAS and its goals.
Question \ref{i:sharing} is future work and it is of interest when resource sharing between different sensor partitions is attempted.

\section{A game-theoretic, doxastic framework}\label{sec:relFrameIR}
Above in \autoref{sec:designproc}, we describe when in the design process of a safety-critical autonomous system \HAS our approach of determining relevance can be helpful. 
In this section we are concerned with the formal ingredients of the framework. 
We discuss  the decisions taken in the design of the framework and point to related work.

In the terms of the IR literature, our relevance notion can be described as 
\emph{situational} --the circumstances of \HAS are taken into account--, 
\emph{subjective} --relevance is determined from the view point of \HAS--, 
\emph{goal-implied} --the goals of \HAS determine whether \HAS misses something relevant--, 
\emph{temporal and spacial} --the performance of \HAS during a maneuver is examined within space and time as captured in \universeD.
The framework integrates these different dimensions, so that we can apply game theory to determine what observations and knowledge is necessary.

\emph{How does the framework integrate so many dimensions of relevance? How can a decision-procedure answer whether something is relevant?}

In short, we model beliefs on the one hand and we use a model of the application domain, \universeD, as ground truth on the other hand. 
We link the two via a two-player dynamic game -- one player is the autonomous system \HAS and the other player is the environment.	

%
\subsection{Scope of the Framework}
%
We aim to support the development of safety-critical autonomous systems that can partially observe their environment. 
Their perceptions may be perturbed or may be contradicting each other. 
We assume that a system \HAS additionally uses its knowledge base to construct its beliefs. 
The knowledge base holds insights about the application domain, that an engineer provided at design time, as well as statements that \HAS gets from trusted sources during its mission.

By asking \enquote{What knowledge and what observations are necessary to build beliefs upon which \HAS can achieve its goals?} we treat belief formation as a central ingredient of our framework. 
Accordingly we use a \emph{doxastic model}, that is a model that captures beliefs explicitly.

\begin{figure}[htbp]
	\centering
	\includegraphics[width=0.5\textwidth]{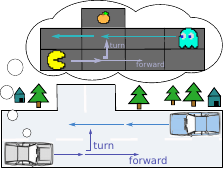}
	\caption{The beliefs of system \HAS can substantially vary from the ground truth world \universeD}\label{fig:beliefDeviation}
\end{figure}

A system \HAS necessarily builds approximating beliefs since its environment is vastly complex while its resources are limited (cf. \autoref{fig:beliefDeviation}).
A system \HAS aims for beliefs that capture the \emph{relevant} aspects.
Allowing the most freedom in building such beliefs provides the greatest potential for saving resources. 

A belief describes what a system \HAS thinks is currently possible. 
To this end we use the possible world semantics \cite{reasoningA}. Accordingly, a belief is a set of possible worlds. 
Since our worlds capture their believed history, current state and future  we call them alternative realities (cf. \autoref{fig:relevanceRelationHAS}).%
\footnote{The possible worlds semantics is often captured via Kripke structure $K$ where $K$'s states represent the worlds and $K$'s state transitions represent the accessibility relation, \ie $s\rightarrow s'$ means in world $s$ $s'$ is a possible world.},
We model that \HAS judges the best action based on its current belief. It does this by simulating whether the action will lead to a mission success in the future  of the believed realities.

\begin{figure}[htbp]
	\centering
	\includegraphics[width=0.7\textwidth]{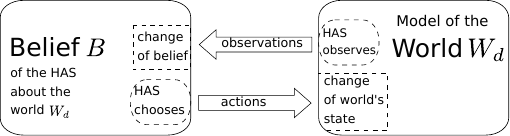}
	\caption{The beliefs and the design time world \universeD are linked.}\label{fig:worlds}
\end{figure}

Since we aim to characterize whether the system \HAS achieves its goals when choosing its actions based on its believes, we link the belief formation to ground truth \universeD, as illustrated in  \autoref{fig:worlds}. 
The feedback loop of \enquote{A system \HAS builds its beliefs based on its perceptions of \universeD.},
\enquote{A system \HAS chooses its actions based on its beliefs.} and
\enquote{A system \HAS's actions influence the state of \universeD.} establishes this link.

Since we want to determine whether a sufficient belief can be formed by approximation of the ground truth \universeD,  we explicitly support beliefs that are structurally distinct from \universeD (cf. \autoref{fig:beliefDeviation}).
Therefore, the ground truth \universeD and the beliefs of \HAS are two separate structures in our framework.

In the framework, we model that \HAS has a knowledge base that captures insights built-in by engineers or received by trusted sources. 
The entries of the knowledge base represent \emph{believed knowledge}, that is \HAS thinks that the entries are true. 
But it is possible that the statements are false.
Our motivation of modelling a base of believed knowledge is, that \HAS will be equipped with rules approximating the reality. 
In order to detect rules and insights that are too coarse, we have to be able to model them in our framework. 
In the sequel we will often refer to the entries of the knowledge base simply as knowledge.

Given a belief formation, we use game theory to determine whether an autonomous safety-critical system \HAS will be successful in \universeD. We also use game theory to determine whether \HAS can form beliefs such that it will be successful. 
We regard a maneuver of \HAS as a dynamic game of the player \HAS and the surrounding world, which might include other agents. 
The system \HAS can control its actions while concurrently the environment chooses from its actions.  The combined actions determine the state change of the \universeD. 
We hence can examine evolutions along \HAS's maneuver in time and space with evolving context.

At its core relevance is a relationship, as mentioned in \autoref{sec:relIR}. 
We examine what knowledge and observations of the world are relevant for \HAS.
\enquote{$X$ is relevant} entails \enquote{having made observation $X$/knowing $X$ makes a difference to \HAS} and not knowing/observing $X$ would hinder \HAS in achieving its goals in \universeD. 
We capture this aspect by defining knowledge/observations $X$ to be relevant, if there is no \enquote{smaller} $X'$ which enables \HAS achieving its goal (cf.~\autoref{def:wr}). 
We do this analogously to \cite{DF11}, where the minimal perimeter of a world model is determined. 
In a nutshell, we explore how well a system \HAS performs when we omit knowledge and observations.
If omission leads to a worse performance the omitted was relevant.

The framework is concerned with \HAS's assessment of the environment via its sensors. 
To this end it only models first-order beliefs but not higher-order beliefs, i.e., beliefs about beliefs. 
Thus system $A$ cannot argue: \enquote{System $B$ will slow down -- I think that $B$ thinks there is a speed limit} or \enquote{System $B$ will slow down -- I  think that $B$ thinks that I think that $B$ should slow down}.  
Including higher-order beliefs will increase the overall complexity of the model. 
There are certainly application where modelling high-order beliefs is essential. We imagine that higher-order beliefs are essential when designing entertaining or comfort functions. 
There the mental state of a user has be taken into account and the system aims to optimally support the user rather guaranteeing goals. 
In contrast, safety-critical systems usually take decisions based on conservative approximations in order to be on the safe side.

In game theory  \emph{rationality} is an central notion. 
Basically, a rational agent $A$ does what promises to result in the outcome $R$ that is best for $A$.
Different notions of rationality exist in literature varying in how to precisely and appropriately capture this notion for a given application.
We assume that \HAS chooses the action that it thinks will lead to the best result.
The system \HAS simulates the effect of its actions in its mind, i.e. it examines the effect on the current set of possible worlds.
So \HAS takes rationally belief-based decisions. 
We do not assume though, that \HAS rationally forms beliefs. 
For instance, we allow  that \HAS believes an object to be red, although according to its observations it is blue, we also allow \HAS to believe that an object is a house at one time instance and at the next time instance \HAS believes it is a tree.
We decided not to constrain the belief formation because of the way beliefs are constructed in autonomous systems. 
The belief of \HAS may be determined by a composition of different components, and there may not necessarily be an entity that ensures that the resulting belief is rational\footnote{What rational in this context should capture, would have to be discussed first.}. 

Our framework, nevertheless, supports the study of different kinds of belief formation functions and we consider it future work. 
We imagine that during the design, requirements regarding the belief formation might be specified. 
So, whether a belief formation exists, that satisfies the requirements, might be valuable insight when developing safety-critical autonomous systems. 
In this line of research, we are also interested in the formalisation of classes of requirements on the belief formation.
In particular, we are interested in belief formations satisfying certain robustness or stability criteria. 
A notion of robustness of  belief formation might express that a given rate of object misclassification can be tolerated.
A stability criterion might express that the beliefs are formed such that replanning is rare and triggered sufficiently early.

%
\subsection{Works related to the Formal Approach \cite{doxFrame}}
%
Epistemology is the theory of knowledge and concerned with information-processing and cognitive success \cite{CollinsEpi,Sheffield}.
Doxastic means \enquote{relating to belief}~\cite{Collins}. 
By using the term \enquote{doxastic}, we want to stress that our formalism focuses on beliefs.
In the epistemic logic literature, the semantics of doxastic languages are often given via \emph{doxastic models}, that are special Kripke structures \cite{EpistemicsLogics}.
A doxastic model $(S,v,\rightarrow_i)$ consists of a set of nodes $S$ representing possible worlds $w$, a valuation function $v: S\rightarrow 2^{\Props}$ for the set of atomic facts \Props and a belief relation $\rightarrow_i$  for each player $i$, that specifies \enquote{$i$ deems $w'$ possible in $w$} if \enquote{$w$ $\rightarrow_i$ $w'$}. With other words, the belief of $i$ at $w$ is defined as the worlds accessible via the agent $i$'s belief relation, $\rightarrow_i$ \cite{EpistemicsLogics,DoxasticMeyer2003}. 

In this paper, we use complex possible worlds instead of the plain nodes of a Kripke structure. 
In our framework, each possible world is a Kripke structure itself, called alternative reality.
It encodes the believed histories, the current states and possible futures. 
A system \HAS uses alternative realities to simulate its strategy in order to decide on its current action. 
In our framework, a reality constitutes an extensive form two-player zero-sum game, where the winning condition is defined by the list of linear temporal logic (LTL) goals of \HAS.
The belief formation is based on partial observations and the currently available knowledge/strong beliefs.

A couple of epistemic temporal logics have been suggested for specifying aspects of knowledge throughout time for multi-agent systems. These logics combine temporal logics with knowledge operators, like KCTL \cite{CTLKMC}, KCTL$^*$ or HyperCTL$^{*}_{lp}$~\cite{HyperCTL}. They are interpreted over Kripke structures. 
But since an agent $i$ has its local view, only certain propositions are assumed to be observable, so that an observational equivalence relation $\sim_i$ on the traces arises. 
\enquote{Agent $i$ knows $\varphi$} then means that $\varphi$ holds on all $i$-equivalent initial traces. 
The alternating time temporal logics (ATL)~\cite{ATL} has been developed for reasoning about what agents can achieve by themselves or in groups throughout time. 
In ATL, the path quantifiers of CTL are replaced by modalities that allow to quantify paths in the control of groups of agents.
ATL is interpreted over concurrent game structures~(CGS), which are labelled state transition systems. 
By adding a knowledge operator, ATL has been extended to an epistemic variant, ATEL~\cite{ATEL}. 
To this end the concurrent game is extended by an observational equivalence relation per agent modelling the agent's limited view.

Just like the logics above, we assume that \HAS can only partially observe the ground truth.  
Our beliefs, however, cannot straightforwardly be expressed in terms of an equivalence on the ground truth, since an alternative reality may be a distinct Kripke structure and a belief does not have to include the ground truth.
In contrast to be above logics, we use in our framework a variant of LTL to specify constraints on the beliefs. 
A  so-called  BLTL formula is therefore interpreted on a belief \belief, i.e. a set of alternative realities. 
Since the set of possible beliefs \Beliefs is finite, a formula $\K\varphi$ means the finite conjunction $\bigwedge_{\reality\in\belief}\reality\models\varphi$.

The field of epistemic planning is concerned with computing plans (\enquote{a finite succession of events} \cite{DELStrat}) that achieve the desirable state of knowledge from a given current state of knowledge~\cite{GIntroDEL}.  
DEL, dynamic epistemic logic, is a formalism to describe planning tasks succinctly by a semantic and action model based approach. 
Epistemic models capture the knowledge state of the agents, and epistemic action models describe how these are transformed. 
An evolution results from  a stepwise application of the available actions. 
In \cite{DELStrat} distributed synthesis of observational-strategies for multiplayer games are considered.  
While ATEL and DEL allow for reasoning about a combination of knowledge and strategies, we are interested in the belief \emph{formation}. 
We ask whether there exists a belief formation that justifies a strategy that successfully achieves temporal goals within a given ground truth world. 

Properties of belief formation are studied in the field of belief revision and update. 
Belief revision is done when a new piece of information contradicts the current information, and it aims to determine a consistent belief set. 
Belief updates may be necessary when the world is dynamic~\cite{BelRevIntro}. 
The works in this field are concerned with rational belief formation, following e.g. some guiding principle like making minimal changes~\cite{BelRevIntro}. 
In our work, we consider very general belief formation functions, since we focus on safety-critical autonomous systems.

BDI agents are rational agents with the mental attitudes of belief~(B), desire~(D) and intention~(I)~\cite{BDI}.
Beliefs describe what information the agent has, 
desires represent the agent's motivational state and specify what the agent would like to achieve, 
while intentions represent the currently chosen course of action. 
These attitudes allow an agent balancing between deliberation about its course of action and its commitment to the chosen course of action.
In our framework, an agent deliberates about its course of action at each state. 
We do not enforce commitment to a certain course of action, as we are interested in whether some belief formation exists.
Nevertheless, the framework conceptually allows capturing notions of commitment, and we plan to examine these in future work. 
Basically, a chosen action represents a set of believed best possible world strategies. These can be considered as the current intent. 
So, a notion of commitment could require that (some) strategies of the previous belief are still best strategies in the current belief.
An engineer may then specify when a system should be committed.

\section{Ingredients of our Doxastic Framework \cite{doxFrame}}\label{sec:ingredients}
\normalsize
In this section, we introduce the ingredients of our framework alongside a running example.
The section is taken from \cite{doxFrame} and slightly enriched  (e.g. by \autoref{ex:Bel}  and \autoref{ex:lin}).

\begin{figure}[htbp]
	\centering
	\includegraphics[width=.5\textwidth]{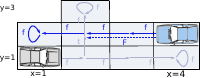}\\
	\caption{Sketch of a simple world}	
	\label{fig:setup}
\end{figure}

We consider two cars, \ego and \other, that are on separate lanes heading towards each other.
The left car, \ego, is our autonomous system \HAS. 
Its goals are avoiding collisions and to take the left turn.
From \ego's perspective \other is uncontrolled.
\autoref{fig:setup} sketches the initial setup and the possible actions of the two cars.  

To formally describe what the two cars can do and how the initial situation may evolve in time and space, we use a labelled Kripke structure as defined in Def.~\ref{def:world} and call it a world. 
\subsection{A World}
In our context, a world models the sphere of interest
and, for this purpose, represents its entities and their actions.\footnote{Thus we use the term \emph{world} here to denote digital or academic worlds, according to \cite{merriamWorld}}
\Ego and other agents are hence part of the world. 
They simultaneously choose actions and thereby how the world transitions from one state to the next. 
We assume that the actions are partitioned into \ego's actions, i.e. the ones that \ego can control, and the ones outside of \ego's control.
\begin{mdefinition}{world}\label{def:world}
	Formally, a \emph{world} \universe is a labelled Kripke structure 
	$\universe=(\States,\Edges,\Label,\Init)$, 
	where 
	\begin{itemize}
		\item \States is the set of states, 	
		\item $\Init\subseteq\States$ is the set of initial states, and  
		\item $\Edges\subseteq\States\times\States$ is the transition relation defining edges between states, 
		\item $\Act=\Act_{\ego}\times\Act_{\env}$ is a finite set of tuples defining the simultaneous actions of our autonomous agent \ego and its environment \env, which may include other agents.	
		\item $\Props$ is  a finite set of atomic propositions 

		\item $\Label=\Label_\States\cup\Label_{\Edges}$ where
			\begin{itemize}
				\item $\Label_\Edges=\Edges\rightarrow {2^{\Act}}$ labels each edge with  a subset of $\Act$ and 
				\item $\Label_\States:\States\rightarrow {2^{\Props}}$ labels state with a subset of \Props,
			\end{itemize}

	\end{itemize}
	We assume that the transition relation is defined for all states and actions, i.e., $\forall \state\in\States,\forall \act\in\Act,\exists \edge\in\Edges: \act\in\Label_\Edges(\edge)$.
\end{mdefinition}
The edge labels $\Label_\Edges(\state)$ of an edge $(\state,\state')$ denote the set of actions the lead from the state \state to state \state'. 
The state labels $\Label_\States(\state)$ denote the set of atomic propositions that are valid at \state. 
We assume that all propositions have a finite domain and hence can be encoded as a finite combination of Booleans. 
In order to express that an action is not enabled at a state \state, \universe can transition into a dedicated state $\sundef$ that is accordingly labelled.\footnote{In this paper any strategy has hence to avoid $\sundef$.}

A sequence of states $\Path=\state_0\state_1\ldots\state_n\in\States^{*}\cup\States^{\omega}$ is a \emph{path} in \universe iff $\forall i, 0\leq i<|\Path|: (\state_{i},\state_{i+1})\in\Edges$.
A path hence describes a possible evolution of world's state. 
$\Path(i)$ denotes  the $i$-th state, $\state_i$. 
$\Path_{< m}$ denotes the prefix of the first $m$ states, $\state_0\ldots\state_{m-1}$, and $\last(\Path)$ is the last state of a finite path $\Path$. $\Path$ is initial iff $\pi(0)\in\Init$. 

Given a tuple $t=(a,\ldots,z)$ we assume that indices carry over to the components, \ie $t_i=(a_i,\ldots,z_i)$. 
\begin{mexample}{A world}\label{ex:uni}
	In our running example, the actions of \ego are \textsf{f}, \enquote{moving one step forward if possible}, \textsf{t}, \enquote{turn and move one step forward}. 
	\Other is either a slow car or a hasty car. If \other is slow it moves one tile forward. 
	If \other is hasty, it leaves its initial position by moving two tiles forward, from all other positions it moves one tile forward. 
	\Other's actions are \textsf{f} and \textsf{F}, \enquote{move two positions forward}. 
	The actions of \ego and \other are annotated by pale blue and dark blue arrows in \autoref{fig:setup}.

	The propositions $\Props_{\textit{pos}}=\Props_{\xego}\cup\Props_{\yego}\cup\Props_{\xother}$ with  $\Props_{\xego}=\{\xego=i|1\leq i\leq 4\}$, $\Props_{\yego}=\{\yego=i|1\leq i \leq 3\}$, $\Props_{\xother}= \{\xother=i|1\leq i \leq 4\}$ encode the positions of the two cars, where $\xego=i$ and $\yego=i$ represent the horizontal and vertical position of \ego, and $\xother=i$ represents the horizontal position of \other. Its vertical position is always two. 
	\Other's car type is encoded via the propositions \textsf{s} (slow) and \textsf{h} (hasty). 
	We assume that \ego cannot observe \other's car type directly, but it has sensors perceiving \other's colour, which is either red or blue. 
	The proposition \blue (\red, \textsf{h}, \textsf{s}) is true, iff \other is a blue (red, hasty, slow) car. 
	The propositions \bp and \rp encode what \ego perceives as \other's colour. 
	They are used to modeling \ego's imperfect colour recognition,  while the propositions \blue and \red encode the true colour of \other. 
	We assume that \ego's colour perception works correctly, when \ego and \other are less than two tiles apart, otherwise the sensor switches colours ($\bp=\neg\blue$, $\rp=\neg\red$). Let $\Props_{\textit{cartype}}$ be the set $\{\textsf{h, s, \blue, \red, \bp, \rp}\}$. 
	The propositions in our example are hence $\Props=\Props_{\textit{pos}}\cup\Props_{\textit{cartype}}\cup\{\textsf{undef}\}$, where \textsf{undef} labels the sink state.

	\begin{figure}[htbp]
		\centering
		\includegraphics[width=0.85\textwidth]{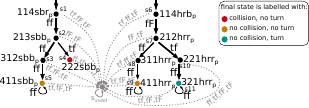}\\
		\caption{Kripke structure of the setup sketched in \autoref{fig:setup}}\label{fig:wm}
	\end{figure}
	Figure \ref{fig:wm} shows the Kripke structure of this world.
	States are labelled with the propositions that hold in the respective state. 
	The label $\mathsf{abcdef}\in\mathbb{N}^3\times\{f,s\}\times\{b,r\}\times\{b_p,r_p\}$  encodes that $\xego=a,\yego=b,\xother=c$ are true and \other's car type is \textsf{d}, its  colour is \textsf{e} whereas the perceived colour is \textsf{f}. 
	The valuations of all other propositions are false. 
	Likewise, the label \textsf{undef} encodes that the only valid proposition is \textsf{undef}.   
	Edges are labelled with sets of actions. We omit the sets' brackets for brevity.
	For example, the label \textsf{ff} denotes the set $\{\textsf{ff}\}$, that contains the one action \textsf{ff}, where \ego and \other simultaneously move one step forward, if possible.   
	Actions that are not enabled at a state lead to the sink state \sundef. 
	We omit the sink state and sink transitions in the sections that follow.
\end{mexample}
In our running example we consider the above world as our \emph{design-time world} \universeD, i.e. we use it as the reference for what is true, as \emph{ground truth}. 
As  discussed in \autoref{sec:designproc}, the design time world \universeD is the result of an analysis activity of the system design. 
\universeD represents the intended application domain including test criteria that the system must master. 
%
%
%
\subsection{Goal List}
Our system \ego has to achieve a prioritized list of goals.
A \emph{goal} \goal is a linear-time temporal logic (LTL) formula~\cite{PrinciplesOfMC}. 
We denote the temporal operator \enquote{globally} by $\Box$, \enquote{eventually} by $\Diamond$, \enquote{next} by \Next and \enquote{until} by \Until. 
We interpret the LTL formulae over (infinite) \emph{traces}, which are infinite
sequences $\cmp = \cmp_0 \cmp_1\ldots \in (2^\Props)^\omega$ of valuations of \Props. 
Satisfaction of an LTL formula $\varphi$ by
a trace $\cmp$ is denoted as $\cmp \models \varphi$.

\begin{mdefinition}{goal list}\label{def:goals}
	A \emph{goal list} $\goalList = (\Goals, \prio)$ consists of a set $\Goals$ of LTL formulae and a priority function $\prio : \Goals \rightarrow \{1, \ldots, |\Goals|\}$ where
	$\goal\in\Goals$ is more important than $\goal'\in\Goals$ iff $\prio(\goal)<\prio(\goal')$.
\end{mdefinition}
We say that a trace $\cmp$ satisfies \goalList with priority $n$ if \cmp satisfies all goals of priority $n$ and more importance, \ie $\cmp \models \goal$ for all $\goal \in \Goals$ with $\prio(\goal) \leq n$.  
A set of traces \Cmp satisfies \goalList with priority  $n$, if all $\cmp\in\Cmp$ satisfy \goalList with priority $n$.
A set of traces \Cmp satisfies \goalList up to priority $n$, if \Cmp satisfies \goalList with priority $n$ and $n$ is the greatest such priority.

For technical reasons the most important goal is $\varphi_g=\true$ and the second most important goal is $\varphi_{u}=\Box \neg \textsf{undef}$.
$\varphi_g$ ensures that at least one goal of the list can be realised. 
$\varphi_{u}$ results from our encoding of disabled transitions: Since we assume that the transition relation is total, we let disabled transitions lead to the state \sundef that is labeled with \udef. 
A strategy is not supposed to take disabled transitions, hence the state \sundef has to be avoided. 
Since we can simply shifting all goals by down-grading their priority and then insert $\varphi_g$ and $\varphi_u$ as the to top most goals, we neglect this issue in the following.

\begin{mexample}{Prioritized Goals}\label{ex:goals}
	We formalize collision freedom as $\varphi_{c}=\Box (\xego=2\land \yego=2\Rightarrow \xother\not=2)$, and $\varphi_{t}=\Diamond (\yego=3)$ expresses that \ego eventually does the turn. 
	The priorities are given by $\prio(\varphi_c)=1, \prio(\varphi_t)=2$.

	Let us now take a closer look at what \ego should do in order to accomplish its goals.
	By inspection of the design time world \universeD, as given  e.g. in \autoref{fig:worldSimple}, we can see that
	if \other is slow, then \ego should not take the turn, but instead it should drive straight on, in order to avoid the collision. 
	If \other is hasty, then  \ego can take the turn and accomplish all its goals.
	\begin{figure}[htbp]
		\centering
		\includegraphics[width=.8\textwidth]{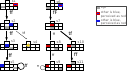}\\
		\caption{Simplified Kripke structure of \universeD where \ego's strategy is highlighted (bold arcs).}	
		\label{fig:worldSimple}
	\end{figure}
\end{mexample}

\subsection{Observations}
\Ego, being highly autonomous, will take decisions based on the beliefs that it has constructed about the world in which it operates. 
\Ego derives its beliefs from the observations made so far and the knowledge/strong beliefs it has about the world. 

A world is usually  only partially perceivable by \ego via observations. \emph{Observations} \Obs are propositions of \universeD whose valuations \ego can assess and that represent  e.g.  sensor readings or received messages from other agents. Observations shed light on \universeD, but they do not have to be truthful, as illustrated in \autoref{ex:uni}, where initially the values of \bp and \rp were switched, so that initially \ego does not perceive the correct colour. 
Despite incorrect observations, a system must, nevertheless, draw conclusions about the current state of the world on the basis of its previous observations. 
We refer to a partial trace leading to a state \state as a history of \state.
\begin{mdefinition}{P-observable history}\label{def:history}
	Let $P$ be a set of propositions, $P\subseteq  \Props$. 
	We call $h$ a ($P$-)\emph{history} of a state \state, if there is an initial path $\Path$ in \universe leading to \state, and
	$h=h_0h_1\ldots h_n$ is the sequence of state labels along $\Path$, $h_i=\Label(\Path(i))\cap P, \forall i, 0\leq i\leq n$. 
	We denote the set of $P$-histories  of \state as $\histories_{P}(\state)$ and the set of all $P$-histories as $\histories_{P}:=\bigcup_{\state\in\States}\histories_{P}(\state)$.  
	We say $h$ is \emph{observable} iff $P\subseteq \Obs$.
\end{mdefinition}
\begin{mexample}{Observable History}\label{ex:hist}
	In our example \ego cannot observe \other's position due to a broken distance sensor, but it can observe its own position and the colour of \other, so $\Obs:=\Props_{x_e}\cup\Props_{y_e}\cup\{\textsf{undef, \bp, \rp}\}$.
	Given the world of \autoref{fig:setup} and \autoref{fig:wm}, \Small{$h\,=\,$114sbr$_p,$213sbb$_p,$312sbb$_p$,411sbb$_p$}%
	\footnote{We apologize for denoting the valuation of \Props rather informal in the following. We do this for the sake of brevity. $114sbr_p$ denotes the valuation where \xego is $1$, \yego is $1$, \xother is $4$, other is a \emph{s}low car, its colour is \emph{b}lue, the \emph{p}erceived colour is \emph{r}ed. Likewise, $11r_p$ denotes that \xego is $1$ and the \emph{p}erceived colour of other is \emph{r}ed.}%
	 is the history along the path \Small{s1$,$s2$,$s3$,$s5} wrt \Props, whereas \ego's observable history wrt \Obs observable history wrt \Obs is \Small{11r$_p$,21b$_p$,31b$_p$,41b$_p$}.
\end{mexample} 
%
%
\subsection{Beliefs} A belief describes what \ego currently thinks the world is like. 
For instance, \ego may think that it saw an approaching vehicle and  that this vehicle is a slow car. 
Due to \ego's belief that the other car is slow, \ego imagines possible future evolutions for a slow car approaching.

We formally capture beliefs as sets of (alternative) realities. 
A reality describes history, current state and possible futures of a world.
\begin{mdefinition}{belief, reality}\label{def:belief}
	A \emph{belief} \belief is the set of realities that \ego currently deems possible, $\belief=\{\reality_0,\ldots,\reality_n\}$. 
	A \emph{reality} is a pair $\reality=(\world,\cStates)$ of a \emph{(possible) world} $\world=(\States,\Edges,\Label,\Init)$ and a set of believed current states $\cStates\subseteq\States$, where any current state is reachable from an initial state and every path has at most one current state.
\end{mdefinition}
A reality specifies a set of current states, that represent the system's assessment of the current state of the world. 
Thereby a reality defines the possible pasts and futures: pasts are captured by the set of paths between initial states \Init and current states \cStates, the possible futures are paths from the current states. 

We also use the term \emph{alternative reality} to stress that a reality is only one possibility that \ego thinks is possible. 
\begin{mexample}{Alternative Realities and Beliefs}\label{ex:altBel}
	To illustrate the notion of belief (cf. Def.~\ref{def:belief}), let us consider the two alternative realities of \ego as illustrated in \autoref{fig:exBel}(a)+(b). The believed past is marked by framing state labels.\\[2mm] 
	\begin{figure}[htbp]
		\begin{minipage}{0.48\textwidth}   
			\centering		
			\includegraphics[width=.4\textwidth]{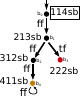}\\
			{\Small{(a) \ego is at $b_0$, \other is slow}}
		\end{minipage}
		\hfill
		\begin{minipage}{0.48\textwidth}
			\centering
			\includegraphics[width=0.4\textwidth]{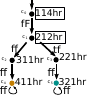}\\
			{\Small{(b) \ego is at $c_1$, \other is hasty.}}
		\end{minipage}
		\caption{Two alternative realities of \ego.		
		The alternative reality of (a) describes that \other is slow and that \ego itself is at the initial state $b_0$ of that world.
		In (b) \other is hasty and \ego is at $c_1$, the \enquote{second} state of that world. 
		The believed history is \Small{\textsf{114hr},\textsf{212hr}} The current states are in bold frames. The history in normal frames.
		}\label{fig:exBel}
	\end{figure}
	Since a belief is a set of alternative realities, singletons of either (a) or (b) form a belief. Also, the set of (a) and (b) forms a belief, where \ego thinks both alternatives are possible.
\end{mexample}
%

\begin{excursus}{Believing in a Different World}\label{ex:Bel}
We want to recall that a system \HAS may believe in worlds that are substantially different from the assumed ground-truth as modelled in \universeD.
An autonomous system \HAS usually captures its application domain by simplified concepts and rules, that reflect its application domain coarsely but sufficiently. 
	Although our examples do not illustrate this point, the framework can be used to form alternative realities that are very different from the design time universe \universeD. 
	\HAS can for instance believe that its actions have a different effect than they have in reality. 
	It can believe it is at situations that are impossible in \universeD, \ie in its beliefs there are valuations of \Props that do not occur in \universeD.

	We assume in this paper though, that\\
	\assumption{act}{the believed actions are a subset of the actions of \universeD}, and
	\assumption{prop}{the propositions in possible worlds are a subset of the propositions \Props.}\\ 
	These two assumptions simplify the framework, but can easily be dropped. 
	But even keeping this restriction is not a severe limitation, since the beliefs do not have to reflect the design time world truthfully. 
\end{excursus}
Since a system \HAS has only finite resources, we assume it can only represent finitely many beliefs, that is, its \emph{set of possible beliefs} $\Beliefs$ is finite.

We write \Small{$\Worlds(\belief)$} for the set of worlds of a belief, \small{$\Worlds(\belief):=\bigcup_{(\cStates,\world)\in \belief}\{\world\}$}. We denote the set of \emph{possible worlds}, \ie the set of worlds taht occur in any possible belief, \Small{$\bigcup_{\belief\in \Beliefs}\Worlds(\belief)$}, as \Small{$\Worlds$}. 

The choice of \Worlds and \Beliefs constitutes an important design decision within the development process of \HAS, as it delimits the expressive power of beliefs. 

\begin{mexample}{Possible Beliefs \Beliefs}\label{ex:bel}
	Our \ego has been designed to represent a certain set of scenarios, for which it can evaluate what to do by extrapolating the future. 
	Figures \ref{fig:belSetup}(a)-(d) sketch a set of possible worlds. 
	The other car may be hasty or slow, the road may be up to 6 tiles long, the intersection may be at \Small{$x=2$} or \Small{$x=3$}, and the start position of \other varies from \Small{$x=4$} to \Small{$x=6$}.
	Note, that \universeD of \autoref{fig:wm} is described by \autoref{fig:belSetup}~(a). 
	\begin{figure}[htbp]
		\addtocounter{figure}{1}    
		\begin{minipage}{0.34\textwidth}
			\centering
			\includegraphics[width=\textwidth]{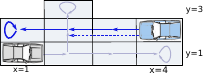}\\
			(a)
		\end{minipage}
		\hfill
		\begin{minipage}{0.38\textwidth}
			\centering
			\hspace*{-13mm}\includegraphics[width=\textwidth]{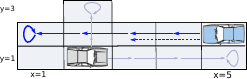}\\
			(b)
		\end{minipage}\\[3mm]
		\begin{minipage}{0.34\textwidth}
			\centering
			\includegraphics[width=\textwidth]{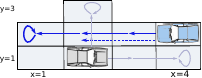}\\
			(c)
		\end{minipage}	
		\hfill
		\begin{minipage}{0.45\textwidth}
			\centering
			\includegraphics[width=\textwidth]{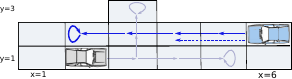}\\
			\hspace*{12mm}(d)
		\end{minipage}
		\addtocounter{figure}{-1}    
		\caption{Sketch of \ego's possible beliefs. If \other is hasty, it uses the dashed arrow at first and then the solid arrows. If it is slow, it uses the solid arrow.} \label{fig:belSetup}
	\end{figure}	
	Let \ego's possible beliefs \Beliefs be the beliefs that canonically evolve from these initial scenarios. 
\end{mexample}
%
%
\subsection{Knowledge Base} 
\emph{Believed knowledge} are statements that the system \HAS believes are true.
These can be built-in or they can be provided at some point in time during \ego's mission. 
We imagine that an engineer equips a system during its design with general statements about the application domain (e.g. \emph{\enquote{Cars drive on the street not under.}}, \emph{\enquote{Velocity is the change in position}}).
Moreover we imagine, that during \HAS's mission certain trusted sources (\emph{like a traffic control system}) provide statements that become strong beliefs. Examples of such statements are \emph{\enquote{In settled areas the speed limit is \SI{50}{km/h}}},  \emph{\enquote{I will be on the highway for the next 20 mins.}} or rules like \emph{\enquote{If A promises to give way, I can rely on it.}}. 

We specify the statements that a system \HAS believes in via an LTL variant, which we call \emph{Belief-LTL~(BLTL)}. A BLTL formula can be satisfied by a belief \belief.

\begin{mdefinition}{BLTL}\label{def:bltl}
	
	Syntax:\\
	A BLTL formula is defined via the following grammar: $\K\psi\,|\,\Kc\psi\,|\,\neg\varphi\,|\,\varphi\land\varphi'$, where $\psi$ is an LTL formula and  $\varphi,\varphi'$ are BLTL formulae.\\
	
	Semantics:\\
	A belief \belief satisfies $\K\psi$, \ie $\belief\models\K\psi$, iff $\psi$ is satisfied by all worlds of all alternative realities of \belief, \ie $\psi$ holds on all traces arising at any initial state of any world $\world\in\Worlds(\belief)$.

	$\Kc$ is analogously defined but on traces from the current states. 
\end{mdefinition}

	$\K\psi$ reads as \emph{\enquote{\ego believes to know that initially $\psi$ held}}. 
$\Kc\varphi$ reads as \emph{\enquote{\ego believes to know that currently $\psi$ holds}}. 
Via BLTL we can specify statements about a belief \belief. 
We can describe the believed past using \K, \eg \enquote{I believe to know that at the start of the maneuver the other car was red}. 
Via $\K^c$ we  can describe what the believed current state is, \eg \enquote{I believe to know that now the car is blue} and we can  refer to the believed future \enquote{I believe to know that in future the car will stay blue.} 
Note, that the past, present and future described via a BLTL formula refers to the content of \belief and not to the ground truth. 
Moreover, note, that a BLTL formula does not allow us to specify constraints on the evolution of beliefs.

\begin{excursus}{Linear temporal properties of belief formation}\label{ex:lin}
Note that BLTL does not allow to specify temporal logic properties regarding the belief evolution. Therefore we would need a formalism that is interpreted on sequences of beliefs. We could for instance define LTBLTL:
\begin{mdefinition}{LTBLTL}\label{def:ltbltl}
	Any BLTL formula is an LTBLTL formula. Given $\varphi_1$ and $\varphi_2$ are BLTL formulae $\neg\varphi_1,\varphi_1\land\varphi_2,\Next\varphi_1,\varphi_1\Until\varphi_2$ is also an LTBLTL. 

	The satisfaction relation is defined in the usual way. We hence present here only the \Until case:
	
	An infinite sequence of beliefs, $\bar{\belief}$, satisfies $\varphi_1\Until\varphi_2$, $\bar{\belief}\models\psi$ if there is an $i$ such that $\forall j<i: \bar\belief(j)\bar\belief(j+1)\ldots\models \varphi_1$ and $\bar\belief(i)\bar\belief(i+1)\ldots\models\varphi_2$.
\end{mdefinition}
	LTBLTL formulae would allow to express properties on the belief formed along a run. For example, such a formula could capture (i)+(ii): (i) Initially \ego believes in $\varphi_1$=\enquote{\other is initially blue and stays blue at all times in all alternate realities}. (ii) \Ego continues to believe in $\varphi_1$ until \ego believes in $\varphi_2$=\enquote{\Other is initially red and stays red at all times}.

	We plan to study properties of the belief formation in future work.
\end{excursus} 
We assume that a finite set $\kbase$ of BLTL formulae is given that represents the believed knowledge that a system \HAS can have at any time.
From this, the engineer can select subsets that \HAS at a certain state.  
\kbase represents the knowledge an engineer can equip  \HAS with, \ie the prior knowledge that she establishes 
and the knowledge that can be transmitted to \HAS during its mission.
\begin{mdefinition}{knowledge base}
	A finite set $\kbase$ of BLTL formulae constitutes a \emph{knowledge base} $\kbase\in\kbases$.
A belief \belief satisfies ${\kbase}$, $\belief\models\kbase$, if $\belief\models\varphi$ holds for all $\varphi\in\kbase$.\\
	\Ego's knowledge base varies over time, we hence extend the labelling function $\Label$ by $\LabelK:\States\rightarrow \kbases$ to specify $\kbase$ as the available knowledge base $\kbase=\Label(\state)$ at a state \state. 
Given a history \history, $\kbase_\history$ denotes the knowledge base of $\last(\history)$. 
	\end{mdefinition}
\begin{mexample}{A Knowledge Base}\label{ex:kb}
	Let \ego have the following knowledge base $\kbase=\{\varphi_z,\varphi_t,\varphi_i,\varphi_{ct}\}$ at all states, where\\[-5mm]
	{\small
\begin{enumerate}\itemindent\indred
		\item { {$\varphi_z=\K \Box ((\neg \xother=5 \land \neg \xother=6) \lor \textsf{undef})$\normalsize}} \\\hfill\textit{(\other is at most at $x=4$)}, \\[\nlred]
		\item {{ $\varphi_t=\K \Box (\neg\xego=2 \Rightarrow (\yego=1 \lor \textsf{undef}))$\normalsize}} \\\textit{(a turn is only possible at $x=2$)},\\[\nlred]
		\item\label{it:start} {{ $\varphi_i=\K\, \xego=1$\\\hfill\textit{(\ego starts at $x=1$)}\normalsize}} and \\[\nlred]
		\item {{ $\varphi_{ct}=\K \Box \bigwedge_{t\in\{\textsf{s, h}\}} (t \Rightarrow \Next (t\lor undef)) \land (\neg t\Rightarrow \Next ( \neg t\lor \textsf{undef}))$\normalsize}}\\\hfill\textit{(the initial car type does not change)}.
\end{enumerate}
Note that BLTL formulae are interpreted on realities and these have designated (maneuver) start states and current states. So \autoref{it:start} expresses that \ego starts its the maneuver at position 1, since by Def.~\ref{def:bltl} traces from initial paths are considered. 
In contrast, the formula $\varphi_c=\K^cx_e=3$ expresses that \ego believes that it currently is at position 3, as for $\K^c$ traces from the current states are considered. 
\normalsize}
\end{mexample}			
\subsection{Belief Formation} %
\Ego updates its beliefs \eg when it gets new information from its sensors, a clock tick or a message from another agent.
The belief formation function \LabelB captures formally how \ego builds its belief. 
	\begin{mdefinition}{belief formation, knowledge-consistent}\label{def:bform}
		The belief formation \LabelB, ${\LabelB}:\histories_{\Obs}\times\kbases\rightarrow \Beliefs$, specifies the belief ${\belief}=\LabelB(\history,\kbase)$ that \sys derives after perceiving a history $\history$ of observations $\Obs\subseteq\Props$ and while believing in $\kbase$. 

		A belief formation $\LabelB$ is called \emph{knowledge-consistent}, if all formed beliefs satisfy the respective knowledge base \kbase, i.e., for all paths $\pi$ of \universeD holds, $\LabelB(\history,\kbase)\models\kbase$ where $\history=\Label_\States(\pi)|_\Obs$ and $\kbase=\LabelK(last(\pi))$.
	\end{mdefinition}

Note, that the knowledge base is a mean to anchor \ego's beliefs to the ground truth.
We could for instance (1) label the states with a knowledge base that reflects the ground truth of a formula $\varphi$. 
Then any knowledge-consistent belief of \ego coincides with the ground truth regarding the evaluation of $\varphi$.
We could also enforce e.g. that (2) the system \HAS forms delayed beliefs, i.e. \HAS believes now in what it had observed two steps before. 

\begin{mdefinition}{history of beliefs, $\LabelBh$ }
	A \emph{belief history} is  a finite sequence of beliefs ${\Small{\bar\belief=\belief_0\belief_1\ldots \belief_n}}$.

	For a history of observations {\Small{$h=h_0h_1\ldots h_n$}} and a history of knowledge bases {\Small{$\khistory=\kbase_0\kbase_1\ldots \kbase_n$}}, we denote by $\LabelBh(h,\khistory)$ the \emph{resulting belief history},
	
	\Small{$\LabelBh(h,\khistory):=\LabelB(h_0,\kbase_0)\LabelB(h_0h_1,\kbase_1)\ldots\LabelB(h_0h_1\ldots h_n,\kbase_n)$}.
	
	\end{mdefinition}
	We write $\LabelB(\history)$ and $\LabelBh(h)$ instead of $\LabelB(\history,\kbase_\history)$ and $\LabelBh(h,\kbase)$, when the knowledgebase is clear from the context.

\begin{figure}[htbp]
	\centering
	\includegraphics[width=.8\textwidth]{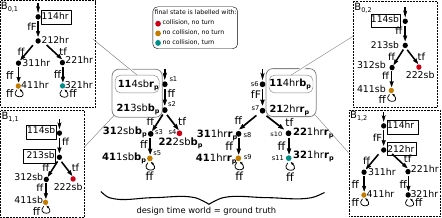}
	\caption{Sketch of a belief formation function}\label{fig:bellab}
\end{figure}

\begin{mexample}{Knowledge-Consistent Belief Formation}\label{ex:knowledgeBelief}
	Let us now consider an example of a knowledge-consistent belief formation.
	We have already defined \ego's possible beliefs in \autoref{ex:bel}. 
	Let \ego have the knowledge base $\kbase':=\kbase\cup\{\varphi_b\}$ at all states, where \kbase is defined in \autoref{ex:kb} and 
	$\varphi_b=\Box (\textsf{h} \Leftrightarrow \rp) \land \Box (\textsf{s} \Leftrightarrow \bp)$ expresses that \ego is also convinced that a red car is hasty, while a blue car is slow.
	In order to satisfy $\kbase$, only realities arising from the scenario (a) of \autoref{fig:belSetup} remain possible.

	 The initial belief of a knowledge-consistent belief formation has to consist of the alternative realities depicted in \autoref{fig:initBel}(a)+(c). 
	\Ego will belief to be in reality $\reality_b$ of Figure~\ref{fig:initBel}(a), when it is in the real world in the scenario of Figure \ref{fig:initBel}(b).
	In this scenario \other is a red car, but \ego incorrectly perceives that \other is blue, hence 	
	$\reality_b$ expresses \ego believe \enquote{I know, \other is blue and slow}. 
	Reality $\reality_b$ moreover describes that \ego thinks to be at the start of the manuever and it captures \ego's expectation of how the future will develop. 
	Similarly, \autoref{fig:initBel}(c) shows the reality $\reality_r$, which \ego things to be in, when it is in the real world in the scenario of \autoref{fig:initBel}(d) where it incorrectly perceives that the blue \other is red.     
	\begin{figure}[htbp]
		\begin{minipage}{0.25\textwidth}
			\centering
			\includegraphics[width=.6\textwidth]{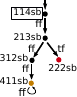}\\
			{(a) \Small{ Alternative reality: Other is slow}}
		\end{minipage}
		\hfill
		\begin{minipage}{0.21\textwidth}
			\centering
			\includegraphics[width=.9\textwidth]{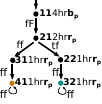}\\[2mm]
			{(b) \Small{Real world: Other is hasty}}
		\end{minipage}
		\hfill
		\begin{minipage}{0.25\textwidth}
			\centering
			\includegraphics[width=.6\textwidth]{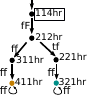}\\
			{(c) \Small{ Alternative reality: Other is hasty}}
		\end{minipage}
		\hfill
		\begin{minipage}{0.21\textwidth}
			\centering
			\includegraphics[width=.9\textwidth]{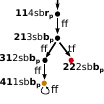}\\[-1mm]
			{(d) \Small{Real world:\\ Other is slow}}
		\end{minipage}
		\caption{(a)+(c): The two alternative realities, where \ego thinks to be at the initial state; the according real world scenarios are (c)+(d) with observable values in bold face type.}\label{fig:initBel}
	\end{figure}	

	So, since \ego's perception is mistaken, \ego is initially convinced that there is a hasty car, when there is a slow car and vice versa.  
	At the initial state, \ego hence thinks that it should do the turn, when it should not. 
	At the next time step, \ego moves one tile forward, while \other simultaneously moves either one or two tiles forward. 
	In both cases, \ego then perceives \other's colour correctly and updates its belief.

	Let us say, \ego considers the more recent observations as more reliable and hence corrects its belief on \other's car type and colour. 
	It updates the belief to \autoref{fig:secondBel}(a) when it is in the scenario 
	\autoref{fig:initBel}(b), and to \autoref{fig:secondBel}(b), when in 
	\autoref{fig:initBel}(d).	
	Figure \ref{fig:bellab} sketches the belief formation so far.
	The observed history, i.e. the tuple of current observations, \Small{11r$_p$} is mapped to belief \Small{B$_{0,1}$} and \Small{11b$_{p} \mapsto $B$_{0,2}$}, \Small{11r$_p$,21b$_p \mapsto $B$_{1,1}$} and \Small{11b$_{p}$,21r$_p \mapsto $B$_{1,2}$}. 

	The sketched belief formation is knowledge-consistent. 
	Note in particular, that in order to be knoweldge consistent, \ego is not required to not change its mind regarding the car type, $\varphi_{ct}$ rather requires that \ego believes that the car type cannot change. That is, $\varphi_{ct}$ has to hold for each formed belief but \ego can form first a belief expressing \other is red and at the next step he can form a belief expressing \other is blue -- \ego would do this in scenario (d) of \autoref{fig:initBel}. 

	\begin{figure}[htbp]
		\centering
		\begin{minipage}{0.6\textwidth}
			\begin{minipage}{0.32\textwidth}
				\centering
				\includegraphics[width=\textwidth]{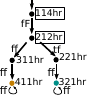}\\
				(a)
			\end{minipage}
			\hfill
			\begin{minipage}{0.29\textwidth}
				\centering
				\includegraphics[width=\textwidth]{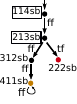}\\
				(b)
			\end{minipage}
			\caption{Alternative realities at the second time step}\label{fig:secondBel}
		\end{minipage}
	\end{figure}	

	Since at the second step, \ego's belief matches the reality, \ego is then able to assess the best strategy matching the real world scenario.
	For its strategic decision \ego can argue along the lines \enquote{Initially I thought the car is red and hasty and that it is a good idea to do the turn. Now I think the car is blue and slow and then the turn is not good idea, since I would collide with other. Since I believe, that my current belief matches the reality, I choose to drive straight on.}  
\end{mexample}

In the above example, all beliefs are singletons, \ie at each point in time \ego believes that there is only one possible reality. The next example illustrates the use of several alternative realities.
\begin{mexample}{Alternative Realities}\label{ex:realities}
	Let us assume that \ego is unsure of its own initial position, thinking that it may initially be at \Small{$x=1$} or \Small{$x=2$} (cf.~\autoref{fig:belSetup} (a)+(c)). 
	So when at state $s1$, \ego deems two realities possible (cf.~\autoref{fig:bellab3});
	in one reality, $\reality_1$, \ego is at \Small{$x=1$}, in the other, $\reality_2$, at \Small{$x=2$}.
	Since we assume that \ego has the same sensors as in \autoref{ex:knowledgeBelief}, in both realities,  $\reality_1$ and $\reality_2$, \other is believed to be red. 
	Similarly, \ego deems two realities possible when at $s6$ (also cf.~\autoref{fig:bellab3}.
	\begin{figure}[h]
		\centering
		\includegraphics[width=.8\textwidth]{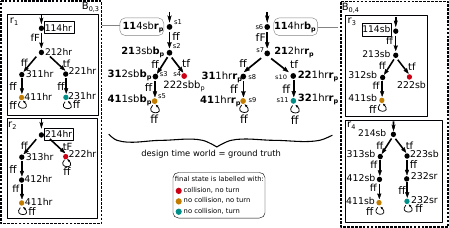}
		\caption{Sketch of a belief formation function when \ego is unsure of its initial position}\label{fig:bellab3}
	\end{figure}
\end{mexample}
%
%
\subsection{Doxastic Model} 
We have by now introduced all components to model how an autonomous system \HAS build its beliefs based an perception and knowledge. 
We summarize the components by defining the notion of \emph{doxastic model}.
\begin{mdefinition}{Doxastic Models} 
	A doxastic model \epi of an autonomous system \HAS is given  by a tuple (\universeD,\goalList,\LabelK,\Obs,\Beliefs,\LabelB) of
	\begin{itemize} 
		\item the design-time world \universeD, 
		\item the prioritized list of goals \goalList, 
		\item the knowledge labelling \LabelK, 
		\item a set of observations $\Obs$, 
		\item the set of possible beliefs \Beliefs of \HAS and 
		\item a belief formation \LabelB.
	\end{itemize}
\end{mdefinition}
The belief formation describes how \HAS links observations made within the world \universeD to its inner representation of the world, i.e. the beliefs \Beliefs that it can possibly build.
The world \universeD is considered as ground truth during the design. 
Later design steps have to take care of the gap between \universeD and the real world.

Note, that we have not yet characterise how the system takes decisions. 
To this end we will introduce the notion of \emph{autonomous decision} (cf. Def.~\ref{def:doxSys}), in the next section that captures that a system takes its decisions based on the its beliefs.

\section{Autonomous Decisions\cite{doxFrame}}\label{sec:autonom}
%
\normalsize
In this section, we formalize a notion of \emph{autonomous decision} and then characterize when a system exists that can take autonomous decisions to accomplish their goals.

Notions of autonomy are discussed in various scientific fields, as we outline in \autoref{sec:autoSys}.
Our formalization aims at capturing that \texttt{(bel\label{as:belief})} an autonomous system must take decisions based on its internal world view (i.e. its belief) and that \texttt{(rat\label{as:rational})} the system chooses the choice alternative that promises the best outcome, i.e. the system is somehow rational. 
We distinguish autonomous decisions from \emph{automatic decisions}, that play out rule-determined choices and are not the rational consequence \wrt the belief content.
The difference between autonomous and automatic decisions is illustrated by \autoref{ex:autonAutom} below.
\begin{mexample}{Autonomous decisions \vs automatic decisions}\label{ex:autonAutom}
	Suppose that \ego has a permanently broken sensor that flips the colors (cf. \autoref{fig:switched}) and that \ego believes in its sensors. 

	\begin{figure}[htbp]
		\centering
		\includegraphics[width=.8\textwidth]{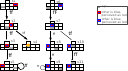}\\
		\caption{Simplified Kripke structure of \universeD: The sensor permanently switches colours. \Ego believes in its sensors and applies the highlighted strategy (bold arcs).}\label{fig:switched}	
	\end{figure}
		So, if \other is red, \ego thinks that \other is blue and vice versa.
	Suppose moreover, that \ego knows that a red car is hasty and a blue car is slow.
	If \ego decides autonomously, then it will follow a strategy highlighted by bold arrows, \ie it will go straight on, if a red car is approaching and it will take the turn when a blue car is approaching. This strategy promises the best outcome \wrt \ego's beliefs.
	 Since it beliefs in its sensors and chooses rationally, it takes the worst possible decisions. 

	Let us now consider a system that plays out automatic decisions. Suppose an engineer is aware, that the sensor switches colours\footnote{Maybe just under certain conditions and hence the engineer did not bother to change the belief formation and decided to patch this problem by implementing a rule}. He hence equips \ego with a rule, that switches the chosen actions accordingly: \enquote{Do not take the turn, when you think a red car is approaching. Takes the turn, when you think a blue car is approaching}. 
		This choice does not make sense to \ego, it is not rational \wrt to its belief content, but, in our scenario it is better when evaluated on the design time world.
\end{mexample}
	The above example \autoref{ex:autonAutom} highlights, that autonomous decisions are not necessarily better than automatic decisions. 
	Since the system \HAS is missing some relevant aspect of the design time world, it takes the wrong decisions. 
	We characterize in Def.~\ref{def:optimal} when a system \HAS can act successfully.
	
\begin{excursus}{Autonomous vs Automatic}
	Although, in the above example the automatic system outperforms the autonomous system,
	autonomous systems promise to be more capable of dealing with new situations. 
	An autonomous system chooses the best possible option based on extrapolation of the system's world view.	
	So once a robust world model for extrapolation has been build, an autonomous system will take \enquote{sensible} decisions.
	Hence a design task when building a system \HAS, is to determine the relevant aspects of the world models and to assess the impact of missing information and sensor perturbations.
	The quality of this extrapolation can be validated by means of runtime monitoring.

Since an automatic system plays out rules, an unforeseen event in the real world might result in situations where no rule applies anymore. 
	The challenge for developing an automatic system hence lies in defining a robust set of rules.  
	This set of rules also has to be evaluated regarding all possible scenarios. 
	It has to be evaluated whether all cases have been identified, when the rule should trigger and whether in these situations, the encoded behaviour is appropriate.
	The expected occurrence a situations satisfying the rule's antecedent are often quite rare, what makes the validation during runtime more difficult.
\end{excursus}

In order to formalize when an \emph{autonomous decision} can be taken successfully, we contrast 
\begin{enumerate}
	\item \emph{truth-observing strategies} (strategies that have access to ground-truth) with 
	\item \emph{doxastic strategies} (strategies that can only observe the formed beliefs) and 
	\item \emph{possible-world strategies} (strategies that run as simulation within the beliefs).
\end{enumerate}
The best truth-observing strategy represents what any system can possibly achieve.
The best doxastic strategy represents what a system with a given belief formation can possibly achieve. 
If the best doxastic strategy performs as good as the best truth-observing strategy, we say that an autonomous system is successful. 
Since our systems choose rationally, they choose what seems to be the best choice according to their belief content. 
The best possible-world strategy describes  what the system believes to be the best strategy in all possible worlds.

The different strategy notions are introduced step by step in the following and an overview of the notions is given in \autoref{tab:nutshell} on page \pageref{tab:nutshell}.
%

\subsection{Truth-Observing Strategy} 
In our framework, we use truth-observing strategies as reference of what would be achievable, if \ego could directly access the ground truth \universeD via a set of propositions $\props\subseteq\Props_{d}$. 
To this end, we say \ego implements a \emph{$\props$-truth-observing strategy} $\strats:(2^{\props})^+\rightarrow\Act_{\ego}$, if \ego chooses its actions based on the history of values of $\props$ as observed in the ground-truth model \universeD. 
When \ego is at state \state of \universeD, a state that was reached via path \Path with $\Label(\Path_{\leq i})|_{\props}=\history$ and $\state=\Path(i)$, it chooses ${\strats}(\history)$. 
A truth-observing strategy \strats together with a sequence of environment actions $\e\in\Act_{\env}^{\omega}$ determines a set of traces, $\Cmp(\e,\strats)$.
Formally, $\Cmp(\e,\strats) = \{\cmp_0 \cmp_1\ldots \in (2^{\Props_D})^\omega |
\exists \textit{ path } \Path \textit{ from } \Init_D, \forall i\geq 0:
\cmp_i = \Label_D(\Path(i)) \land 
\act_i:=\strats(\Label_D(\Path_{\leq i})|_{\props}) \land
(\act_i,\e(i))\in\Label_D(\Path_i,\Path_{i+1})\}$.

\subsection{Doxastic Strategy} Since a system \HAS has no direct access to the ground truth,  it has to decide based on its history of beliefs. 
We formalize this by the notion of doxastic strategy.
At a state $\state=\Path(i)$ in \universeD \ego takes a decision based on the history of its beliefs $\bhist_0\ldots\bhist_i$ that \ego has built along $\Path_{\leq i}$.
So to implement the \emph{doxastic strategy} $\stratb:\Beliefs^{+}\rightarrow\Act_{\ego}$ on $\universeD$,  \ego chooses $\stratb(\LabelB(\Path_{\leq i}))$.
A strategy \stratb together with a  sequence of environment actions $\e\in\Act_{\env}^{\omega}$ determines a set of traces in \universeD, just like for truth-observing strategies. 
The set of traces is $\Cmp(\e,\stratb) = \{\cmp_0 \cmp_1\ldots \in (2^{\Props_D})^\omega |
\exists \textit{ path } \Path \textit{ from } \Init_\epi, \forall i\geq 0:
\cmp_i = \Label_\epi(\Path(i)) \land 
\act_i:=\stratb(\LabelB(\Path_{\leq i})) \land
(\act_i,\e(i))\in\Label_\epi(\Path_i,\Path_{i+1})\}$.

Note that  doxastic strategy indirectly depends on what is observable:  the belief formation \LabelB (cf. Def.~\ref{def:bform}) observes only a certain set of observations.\\

\myparagraph{Dominance, $\strat'\leq_{\universe,\psi}\strat$}Since truth-observing and doxastic strategies both determine traces for a given sequence of environment actions,  we can compare them straight forwardly:  A strategy \strat achieves a goal list \goalList up to $n$ on \universe, if no matter what the environment does, \strat achieves $\goalList$ up to $n$, \ie the set $\bigcup_{\e\in\Act_{\env}^{\omega}}\cmp\in\Cmp(\e,\strat)$ satisfies $\goalList$ up to $n$ (cf. page \pageref{def:goals}).
A strategy \emph{\strat \goalList-dominates a strategy $\strat'$ on $\universe$}, $\strat'\leq_{\universe,\psi}\strat$, iff $\strat'$ achieves \goalList up to $n'$ and \strat up to $n$ where $n'\leq n$.
We also say \emph{$\strat'$ $\varphi$-dominates $\strat$}, $\strat'\leq_{\universe,\varphi}\strat$, for an LTL property $\varphi$, iff $\strat'\leq_{\universe,\psi}\strat$ for the goal list $\psi$ with the singleton goal set $\Phi=\{\varphi\}$.
We omit $\universe$ if it is clear from the context.
\begin{mexample}{Truth-Observing and Doxastic Strategies}\label{ex:tods}
	As an example of a dominant \props-truth-observing strategy, let us consider 
	\begin{itemize}[leftmargin=4mm]
		\item the goal list of \autoref{ex:goals}  on page \pageref{ex:goals}, ($\varphi_c$, \ie no collisions, is more important than $\varphi_t$,\ie do a turn), 
		\item the world model in \autoref{fig:setup}(b) on page \pageref{fig:setup}, 
		\item the propositions $\props:=\{\xego,\textsf{s},\textsf{h}\}$ to be observable by \strats and 
		\item the strategy \strats that chooses to drive straight on, if \other is hasty, and that chooses to turn, if \other is slow\\
	(it maps \Small{$1s\mapsto f$, $1s,2s\mapsto f$, $1s,2s,3s \mapsto f$, $1s,2s,3s,4s\mapsto f$, and  $1h\mapsto f$, $1h,2h\mapsto t$, \ldots}).
	\end{itemize}
	Strategy \strats achieves \goalList only up to $\varphi_c$, \ie collision-freedom, and \strats is a dominant ($\props$-truth-observing) strategy, since in all cases collision-freedom is guaranteed and
	in case the car is slow, no other strategy can do better, \ie realize both, collision-freedom and the turn.

	Let us now consider a doxastic strategy \stratb.
	\begin{itemize}
		\item We consider the same goal list and world model as for \strats.
		\item We take the konwledge-consistent belief formation \LabelB as sketched in \autoref{fig:bellab} on page \pageref{fig:bellab}, \ie \ego always believes in its sensor readings and its sensor initially switches colours.
			
			Its set of observables is $\Obs:=\Props_{x_e}\cup\Props_{y_e}\cup\{\textsf{undef, \bp, \rp}\}$ (cf. \autoref{ex:hist}, p.~\pageref{ex:hist}). The knowledge base is defined on p.~\pageref{ex:kb},~\autoref{ex:kb}, as  $\{$\fs
			$\varphi_z$ (\other is at most at $x=4$),
					{{$\varphi_t$}} (a turn is only possible at $x=2$),
					$\varphi_i$ (\ego starts at $x=1$), 
					$\varphi_{ct}$ (the initial car type does not change)\}\normalsize.

				\item Let \stratb be a doxastic strategy with \Small{$B_{0,1}\mapsto f$, $B_{0,1}\,B_{1,1}\mapsto f$, $\ldots$} and \Small{$B_{0,2}\mapsto f$, $B_{0,2}\,B_{1,2}\mapsto t$, $\ldots$}. \stratb is illustrated in \autoref{fig:belstrat}.				
	\end{itemize}
	\begin{figure}[h]
		\centering
		\includegraphics[width=.98\textwidth]{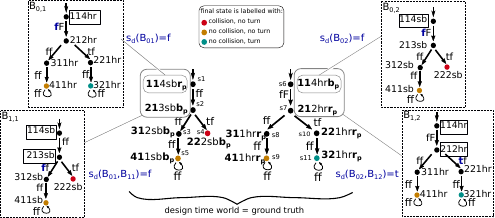}
		\caption{doxastic strategy \stratb; initially the sensor switches colours; \ego believes in its sensor.}\label{fig:belstrat}
	\end{figure}

	Just like \strats,  \stratb chooses to turn when \other is hasty (\Small{$B_{0,2}\,B_{1,2}\mapsto t$}), and it chooses to drive straight on, if \other is slow (\Small{$B_{0,1}\,B_{1,1}\mapsto f$}). 
	As there is \enquote{no better} strategy, \stratb is dominant. 
\end{mexample}

Next, we want to capture that \ego chooses its actions based on the \emph{content} of its beliefs. In order to motivate our formalization, let us consider the following example, where \ego does not choose its actions based on its belief content.
\begin{mexample}{Decisions Not Based on the Belief Content}\label{ex:noCon}
	We modify our running example slightly: Let us assume the colour perception is severely broken and permanently switches red to blue and vice versa.	
	In \autoref{fig:bellab2} the changed world model is given along with a belief formation that relies on the colour perception, i.e., if the sensors say the other car is red (blue), then \ego believes the other car is red (blue).
	\begin{figure}[h]
		\centering
		\includegraphics[width=\textwidth]{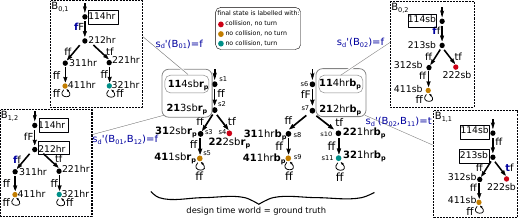}
		\caption{doxastic strategy $\stratb'$; the sensor switches colours permanently; \ego believes in its sensor.}\label{fig:bellab2}
	\end{figure}

	Let $\stratb'$ be a doxastic strategy with \Small{$B_{0,1}\mapsto f$, $B_{0,1},B_{1,2}\mapsto f$, $\ldots$} and \Small{$B_{0,2}\mapsto f$, $B_{0,2},B_{1,1}\mapsto t$, $\ldots$}. 
	Just as \strats and \stratb,  the strategy $\stratb'$ realizes a turn on \universeD, if \other is hasty (\Small{$B_{0,2},B_{1,1}\mapsto t$}), and \ego drives straight on, if \other is slow (\Small{$B_{0,1},B_{1,2}\mapsto f$}). 
	So $\stratb'$ is a dominant strategy, but $\stratb'$ makes no sense from \ego's perspective. 
	In case \other is hasty, \ego believes that \other is slow, since it trusts its sensors and \ego extrapolates that doing the turn  would cause a collision. 
	But in this case, $\stratb'$ demands to take the turn.
	Vice versa,  $\stratb'$ chooses to drive straight on, when \ego believes \other is hasty and it extrapolates that taking the turn is alright. 
\end{mexample}
%
\subsection{Possible-worlds Strategy}
\autoref{ex:noCon} motivates, what it means that \ego decides based on the content of its belief. 
We will formalize this as \enquote{\Ego always chooses an action, that a dominant strategy in \ego's current belief \belief would also choose at the believed current state}.
To capture this formally, we introduce the notion of \emph{possible-worlds strategy}. 

A \emph{possible-worlds strategy}  is a function $\stratw:{(2^{\Props_\belief})^{+}}\rightarrow\Act_\ego$ and it is applied to the alternative realities of \ego's current belief \belief. 
This results in believed traces.
We define this set of traces in an alternative reality $\reality=(\world,\cStates)\in\belief$ for a (believed) sequence of environment actions $\e\in\Act_{\env}^{\omega}(\world)$ as 
$\Cmp(\e,\stratw,\reality) = \{\cmp_0 \cmp_1\ldots \in (2^{\Props})^\omega | \exists \textit{ path } \Path\textit{ in }\world\textit{ from }\Init:  \forall i\geq 0:
\cmp_i = \Labelw(\Path(i)) \land 
\act_i:=\stratw(\Labelw(\Path_{\leq i})) \land
(\act_i,e(i))\in\Labelw({\Path}_{i},\Path_{i+1}) \}$.
We generalize the notion of \goalList-dominance to possible-worlds strategies. 
A possible-worlds strategy \stratw \goalList-dominates a possible-worlds strategy $\stratw'$ in \belief, if \stratw \goalList-dominates $\stratw'$ in all realities $\reality\in\belief$.
\begin{mexample}{Possible-Worlds Strategy}\label{ex:pws}
	Consider the possible-worlds strategy \stratw  that chooses to turn, if \other is hasty, and to drive straight on, if \other is slow, i.e., we consider \stratw with  \Small{$114hr\mapsto f$, $114hr, 212hr\mapsto t$,  $114hr, 212hr,221hr\mapsto f$, \ldots, $114sb\mapsto f$, $114sb,213sbb_p\mapsto f$, \ldots}. \stratw is sketched for the case that \other is hasty via bold arcs in \autoref{fig:bels}. 
	Note that \autoref{fig:bels} shows the excerpts of $B_{0,1}$ and $B_{1,2}$ as in \eg \autoref{fig:bellab2}.  
	$B_{0,1}$  expresses that \ego thinks that it is at the initial state \Small{$114hr$}. 
	\Ego follows \stratw by choosing \Small{$\stratb(114hr)=f$} when having this belief. 
	\begin{figure}[htbp]
		\begin{minipage}{\textwidth}
			\begin{minipage}{0.45\textwidth}
		a)\\[-3mm]
		\hspace*{7mm}\null\includegraphics[width=0.38\textwidth]{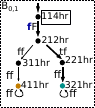}
		\end{minipage}
			\hfill
		\begin{minipage}{0.45\textwidth}
		b)\\[-3mm]\hspace*{7mm}\includegraphics[width=0.38\textwidth]{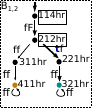}
		\end{minipage}
	\end{minipage}
		\caption{a) Belief $B_{0,1}$ and (b) belief $B_{1,2}$ of \autoref{fig:bellab2}.}\label{fig:bels}
	\end{figure}
	The belief $B_{1,2}$ (cf. \autoref{fig:bels} b) captures that \ego thinks to have already made one move and is now at state \Small{$212hr$}. According to \stratw, \ego has to choose $t$, since \Small{$\stratw(114hr,212hr)=t$}. \Ego hence has to choose $t$ when currently having the belief $B_{1,2}$.
\end{mexample}

\begin{table}
	\FramedBox{6.8cm}{\textwidth}{
		\small
		Strategy tyes:\\[\nlred]
		\begin{itemize}
			\item\small \emph{truth-observing strategy} $\strats:(2^{\props})^+\rightarrow\Act_{\ego}$\\ 
				observes the  ground truth world $\universeD$ via $\props\subseteq\Props_{\universeD}$ and takes decisions based on their history; serves as comparative reference of what is achievable given $\props$ could be observed directly\\[\nlred]
			\item\small \emph{doxastic strategy} $\stratb:\Beliefs^{+}\rightarrow\Act_{\ego}$\\
				observes the beliefs to take decisions and takes decisions based on their history; represents the decision making of autonomous and automatic systems\\[\nlred]
			\item\small \emph{possible-worlds strategy} $\stratw:{(2^{\Props_\belief})^{+}}\rightarrow\Act_\ego$\\
				captures how a system \HAS \enquote{simulates} its strategies within the alternative realities; decisions are taken based on the believed history within the respective alternative reality\\[\nlred]
		\end{itemize}
		A strategy \strat  \emph{\goalList-dominates} $\strat'$, $\strat'\leq\strat$, iff $\strat'$ achieves the goal list \goalList up to priority $m'$ but \strat achieves \goalList up to priority $m'$ with  $m'\leq m$.\\
		\normalsize
		}
		\caption{Strategy types \& dominance in a nutshell}\label{tab:nutshell}
\end{table}

A dominant possible-worlds strategy determines what is the best to do, given a  belief. 
So in order to express that \HAS chooses the action, that it thinks is currently the best, we refer to what a dominant possible-worlds strategy would choose for a given belief.

A peculiarity of possible-worlds strategies is, that they can be \emph{indecisive} for a belief \belief. 
That is, a possible-worlds strategy might determine two or more different actions for the set of believed current states.
More precisely,  \stratw is called current-state indecisive, if there are two paths, $\Path_1, \Path_2$, in \belief leading to believed current states and if \stratw chooses the action $\act_1$ at $\Path_1$ while it chooses $\act_2$ at $\Path_2$:  
\begin{mdefinition}{current-state (in)decisive}
	We call a possible-worlds strategy \stratw \emph{current-state indecisive} in belief \belief iff $\exists\reality_1,\reality_2\in\belief\land \bigwedge_{i\in\{1,2\}}\exists_i\Path_i\in\Paths(\reality_i):\bigwedge_{i\in\{1,2\}}\last(\Path_i)\in\cStates(\reality_{i})\land\stratw(\Label_{\reality_1}(\Path_1))\not=\stratw(\Label_{\reality_2}(\Path_2))$.

	\stratw is \emph{current-state decisive} in \belief iff it is not current-state indecisive in \belief.
\end{mdefinition}

The indecisiveness may result from uncertainties of \ego. 
\Ego might be missing information that would allow it to determine the current situation sufficiently.
Since this information is missing, \ego instead forms a belief with a multitude of realities. 
That way a belief can encode even contradictory information. 
\begin{mexample}{Lack of information and indecisiveness}\label{ex:unknownGoal}
Let us assume that \ego has to get to a filling station on the shortest possible route. 
It is currently not sure where the filling station is.
It hence forms a belief $\belief$ of two realities, $\reality_1$ and $\reality_2$.
In reality $\reality_1$ the filling station is to its left, while in $\reality_2$ the filling station is to its right. 
In $\reality_1$ \ego must to move left while it must move right in $\reality_2$. 
Since \ego deems both realities possible, it cannot decide whether to turn right or left. 
\end{mexample}
\begin{mexample}{Indecisive possible-worlds strategy}\label{ex:undec}
Another example where a possible-world strategy is not able to determine a unique best choice, is given by $B_2$, the belief depicted in \autoref{fig:undec}. 
	\begin{figure}[htbp]
		\centering
		\includegraphics[width=0.5\textwidth]{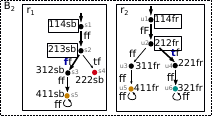}
		\caption{Belief $B_{2}$ describes that \ego believes that \other can be blue or red, and it believes to have made already one step.}\label{fig:undec}
	\end{figure}	
	$B_2$ may be formed because \ego's sensor does not give any information about \other's colour, so that \ego believes that both colors are possible. 
	$B_2$ moreover captures that \ego believes to have made one step, i.e., it believes to be in state $s_2$ of reality $r_1$ or in state $u_2$ of reality $r_2$. 
	The strategy \stratw determines $f$ as the best option due to $s_2$ and $t$ as the best option due to $u_2$ of $r_2$. 
	Hence it is not obvious, whether to choose at the current state $t$ or $f$, given $B_2$.
\end{mexample}
We call the set of actions that a possible-worlds strategy chooses at the set of current states, its current-state choices, $\cAct(\stratw,\belief)$:
\begin{mdefinition}{current-state choices of \stratw, \Small{$\cAct(\stratw,\belief)$} }\label{def:cdec}
	Let \Small{$\cAct(\stratw,\belief)$} be the set with

	$\act \in \cAct(\stratw,\belief)\Leftrightarrow \exists \reality=(\world,\cStates)\in\belief: \exists \textit{ path }\pi\textit{ in }\world: \pi(0)\in\Init\land\textit{last}(\pi)\in\cStates\land\stratw(\pi)=\act$.
	
	We call \Small{$\cAct(\stratw,\belief)$} the current-state choices of \stratw. 
\end{mdefinition}
Given a strategy is decisive at the set of current states, we call it current-state decisive:
\begin{mdefinition}{current-state decisive possible-worlds strategy}\label{def:cdec}
	A possible-worlds strategy \stratw is \emph{current-state decisive in a belief \belief}, if \Small{$\cAct(\stratw,\belief)$} is a singleton.
\end{mdefinition}
Note, that the examples \autoref{ex:unknownGoal} and \autoref{ex:undec} illustrate that the existence of a current-state decisive possible-worlds strategy is not guaranteed.
\begin{mproposition}{Existence of a Current-state Decisive Strategy}\label{prop:ex-csds}
	We can decide whether there is a current-state decisive possible-worlds strategy \stratw achieving an LTL property $\psi$ in a given belief \belief.
	If it exists, we can synthesize such a strategy.
\end{mproposition}
\begin{proofsketch}{Prop.~\ref{prop:ex-csds}}
	We sketch how a current-state decisive possible-worlds strategy for a belief \belief can be synthesized.
	Given a belief \Small{$\belief=\{\reality_1,\ldots,\reality_n\}\in\Beliefs$}, we first build a single reality $\reality_{\belief}$ by the disjoint union of all alternative realities \Small{$\reality_{\belief}:=(($
	$\dbigcup_{r_i}\{\States_i\}$, 
	$\dbigcup_{r_i}\Edges_i$, 
	$\dbigcup_{r_i}\Label_i$,  
	$\dbigcup_{r_i}\Init_i)$, 
	$\dbigcup_{r_i}\{{\cStates}_i\})$}.
	If necessary, we can make the realities disjoint by renaming their states but keeping their structure.

 	We iterate through the list of \ego's actions.
	In iteration $i$, we modify $\reality_{\belief}$ to create the $\reality_i$ where the current action $\act_i$ becomes the only possible choice at all current states $\state_c$. 
	More precisely, in $\reality_i$ all transition that orginate at a current state $\state_c$ and that are labelled with an action $\act\not=\act_i$ are (re)directed to lead to $\sundef$.
	Using \cite{LTLSynth}, we synthesize a winning strategy \strat for $\psi \land \Box \neg \sundef$\footnote{The complexity is 2EXPTIME-complete}. If it exists, we stop iterating. 
	The synthesized winning strategy \strat applyies the same action at all current states, by construction. 
	It obiviously is also a winning strategy of \belief and current-state decisive.
	Since we check for all actions, whether such a strategy \strat exists, the algorithm is guaranteed to find a current-state decisive possible-worlds strategy, if it exists. Since there are only finitely many actions, the algoirithm terminates.
\end{proofsketch}
We consider actions that are the current-state choices of a dominant possible-worlds strategy as  rationally justified choices of an autonomous system. We therefore define the set of current-state choices of a belief:
\begin{mdefinition}{best choices in \belief, \Small{$\bAct(\belief)$} }\label{def:choiceb}
	Let a goal list $\goalList$ be given. Let \Small{$\bAct(\belief)$} be the set with
	
	$\act\in\Small{\bAct(\belief)}\Leftrightarrow \exists\tit{$\goalList$-dominant }\stratw\tit{ in }\belief:\act\in\Small{\cAct(\stratw,\belief)}$. 
\\
	We call \Small{$\bAct(\belief)$} the current-state choices in \belief for $\psi$. 
\end{mdefinition}
The following example illustrates that in a belief \belief there can be several dominant current-state decisive possible-worlds strategies.
\begin{mexample}{current-state decisive and multiple action choices}\label{ex:undecMulti}
	Let us assume as in \autoref{ex:unknownGoal} that \ego has to get to a filling station on the shortest possible route. 
Let us assume \ego forms a belief $\belief$ of only one reality, $\reality$.
In reality $\reality$ there is a filling station to its left and to its right. 
	Hence \ego can turn right --let this be strategy $\stratw_1$-- and it can turn left --strategy $\stratw_2$. 
So \ego can choose to turn right or left according to  $\stratw_1$ and $\stratw_2$, respectively, and both strategies are current-state decisive.
\end{mexample}
\begin{mproposition}{Determining the best choices \Small{$\bAct(\belief)$}}\label{prop:actBel}
	Let a goal list $\goalList$ and a belief $\belief$ be given. 
	
	We can determine the set of best choices \Small{$\bAct(\belief)$} for $\goalList$ in \belief.
\end{mproposition}
\begin{proofsketch}{Prop.~\ref{prop:actBel}}	
	According to Def.~\ref{def:choiceb} \Small{$\bAct(\belief)$}  are the actions that are current-state choices of some $\psi$-dominant startegy $\stratw$ in \belief.
	We first determine the maximal $\mn$ of the goal list $\goalList$ that is achievable by any  possible-worlds strategy in \belief. With other words, a possible worlds strategy is dominant, if it achieves \goalList up to \mn.  
	To this end we check whether we can synthesize, \cite{LTLSynth}, a strategy in $\reality_{\belief}$\footnote{$\reality_{\belief}$ is the disjoint union of all alternative realities as defined in the proof of Prop.~\ref{prop:ex-csds} on page \pageref{prop:ex-csds}} that achieves $\psi$ for priority \mn, starting with $\mn=|\goalList|$ down to $\mn=0$.
	We then proceed as in the proof of Prop.~\ref{prop:ex-csds} on page \pageref{prop:ex-csds}, \ie by examining whether there is a possible-worlds strategy that achieves \goalList up to $\mn$ in the modified reality $\reality_i$ for action $i$, but we do not stop as soon as one could be synthesized but instead we examine all actions \Act.
\end{proofsketch}
\subsection{Autonomous Decision}
In this section we develop our notion of an autonomous decision and we define when systems are autonomous-decisive.\\ %

\emph{For the following let a doxastic model $\epi=(\universeD,\goalList, \LabelK,\Obs, \Beliefs, \LabelB)$ of a system \HAS be given.
	Let \Path be a finite initial path in \universeD, \history the observed history along $\Path$ and $\belief=\LabelB(\history)$ the formed belief.}\\

We call a system that decides based on its beliefs a \emph{doxastic system}.
Its decisions are determined by a doxastic strategy $\stratb$. 
\begin{mdefinition}{doxastic system}\label{def:doxSys}
	A \emph{doxastic system} $\sys$ is a pair $\sys=(\epi,\stratb)$ of a doxastic model $\epi$ and a doxastic strategy $\stratb$, $\stratb:\Beliefs^{+}\rightarrow\Act_{\ego}$ on $\universeD$.  
	For all finite paths \Path in \universeD, the system chooses $\stratb(\LabelB(\Path))$.
\end{mdefinition}
Doxastic systems do not base their decisions on the ground truth or on observations, but on their beliefs.
It is not constrained how they come to a decision though. 
Hence doxastic systems can be \eg automatic or autonomous systems (\cf \autoref{ex:autonAutom}). 

We regard autonomous systems as special doxastic systems, whose decisions are rational \wrt the content of the current belief.
So, $\stratb$ should choose actions that are the current-state choices of a dominant possible-worlds strategy \stratw.
Moreover, we require that \stratw should be current-state decisive.
When no current-state decisive strategy exists, this means that there is  no way to rationally avoid an unwanted consequence.
Then the current-state choice set \cAct(\stratw,\belief) means a gamble: $\act_1\in\cAct(\stratw,\belief)$ might achieve the targeted goal or another action $\act_2\in\cAct(\stratw,\belief), \act_1\not=\act_2,$ would be the right choice.

We hence regard it a design goal to develop systems that form beliefs where a current-state decisive possible-worlds strategy exists -- with other words, we strive to build a system \HAS that always builds a belief where it can determine a choice achieving its goals.
If a belief \belief is formed where no current-state decisive possible-worlds strategy exists, an engineer can adjusting the system's goals (\eg by weakening the goals to \enquote{if you are uncertain, choose the safe option}) or by improving the formed beliefs -- adding additional knowledge or adding sensors/observables.\\
\assumption{csdec}{In the following we assume that the belief formation \LabelB forms only beliefs \belief in which a current-state decisive strategy exists.}\\

A system that is not autonomous-decisive cannot rationally determine which action is currently appropriate. A goal for the  design of \HAS is hence to ensure that a system is autonomous-decisive.

To summarize, we call a decision autonomous, if it is the rational choice for the current belief, that is 
 $\stratb$ chooses actions that are the current-state choices of a dominant, current-state decisive possible-worlds strategy:
\begin{mdefinition}{autonomous decision}\label{def:autonomDec}
	The system \HAS decides autonomously at \Path,  if it chooses an action $\act\in\bAct(\belief)$.
\end{mdefinition}
A system that always decides autonomously, follows a special doxastic strategy $\stratbb$ that always chooses an action $\act\in\bAct(\belief)$ when on a path \Path in \universeD, where $\belief=\LabelB(\Label_{\States}({\Path})|_{ \Obs},\LabelK(\last(\Path)))$ is the belief formed after the observed history along \Path and while having the believed current knowledge $\LabelK(\last(\Path))$.
Note that such a system follows a memoryless doxastic strategy.
The system's memory is \enquote{shifted} into the beliefs. 
The framework thus can capture how a system \HAS deals with the  finite memory also \wrt encoding the relevant.
\begin{mdefinition}{autonomous strategy}\label{def:autonomDec}
	A doxastic strategy $\stratbb : \Beliefs^{+} \rightarrow \Act$ is called an \emph{autonomous strategy}  iff 
		for all belief histories $\bar\belief\in\Beliefs^+$ it holds that {$\stratbb(\bar\belief)\in\bAct(\last(\bar\belief))$}.
\end{mdefinition}

We say that a system \sys autonomous-decisively achieves the goal list $\goalList$ up to $n$, if it implements an autonomous strategy $\stratbb$, \ie $\sys=(\epi,\stratbb)$, and \stratbb achieves $\psi$ up to $n$.

So far we do not require that an autonomous-decisive system \sys behaves appropriately in a given setting.
It is only guaranteed, that \sys acts rationally \wrt its beliefs. 
Its belief formation does not have to reflect the real world though.
Def. \ref{def:optimal} closes the gap.

By Def.~\ref{def:optimal} we basically enforce the belief formation \LabelB to form beliefs, so that \sys is as successful as the best system with direct access to the ground-truth of the design-time world \universeD.
\begin{mdefinition}{Optimal autonomous-decisive system}\label{def:optimal}
	The autonomous system $\sys=(\universeD,\goalList, \LabelK,\Obs, \Beliefs, \LabelB,\stratbb)$ is an \emph{optimal} autonomous-decisive system, 
	if the autonomous strategy $\stratbb$ is not $\goalList$-dominated by any \Props-truth-observing strategy.
\end{mdefinition}
In the following we are focusing on \emph{optimal} autonomous-decisive systems. 
For brevity we usually just speak of \emph{autonomous systems}.
We will discuss the relation of our notion of optimal autonomous-decisive systems with the notion of autonomous systems in \autoref{sec:autoSys}.

Def.~\ref{def:optimal} requires that the belief formation captures the gist of observations \wrt \sys's goals. 
It is a rather flexible way of constraining the belief formation: \LabelB has to preserve the \emph{relevant aspects} of \universeD. 
A more direct way to anchor beliefs in the ground-truth is given by the knowledge base. 

We say that $\sys$ is an \emph{optimal} autonomous-decisive system, 
	if the autonomous strategy $\stratbb$ is not $\goalList$-dominated by any \Props-truth-observing strategy\footnote{an \Props-truth-observing strategy is based on perfect observations of \universeD, cf. Tab.~\ref{tab:nutshell}}.
$\sys$'s belief formation \LabelB then builds beliefs, such that \sys is as successful as the best system with direct access to the complete ground truth, \universeD.

We can decide for a given doxastic model without belief formation, $\epi^{-}=(\universeD,\goalList,\LabelK,\Obs,\Beliefs,.)$, whether there is a knowledge-consistent belief formation \LabelB and
an autonomous strategy \stratbb, and we can synthesize the two (cf.~\autoref{th:auto}).  
\begin{mtheorem}[Autonomous Decisiveness~\cite{doxFrame}]\label{th:auto}
	Let  $\epi^{-}=(\universeD,\goalList,\LabelK,\Obs,\Beliefs,.)$ be a doxastic model without belief formation.

	We can decide whether there is a knowledge-consistent belief formation \LabelB and a  doxastic strategy \stratb such that $\sys=(\epi^{-},\LabelB,\stratb)$ is an optimal autonomous-decisive system. 
	If such \LabelB and \stratb exist, we can synthesize them. 
\end{mtheorem}
\begin{proofsketch}{\autoref{th:auto}}
	The proof can be sketched as follows. 
	We build a Kripke structure $\universeD'$ such that any \Obs-truth-observing strategy \strats in $\universeD'$ encodes (i) a belief formation $\LabelB$ and (ii) an autonomous strategy \stratbb, such that (a) \LabelB is knowledge-consistent and (b) if \strats achieves \goalList up to $n$, also $\stratbb$ does. 
	The idea for the construction of $\universeD'$ is as follows. In $\universeD'$ the strategy \strats does not choose actions but beliefs.	
	Therefore,  the transitions in \universeD are copied to $\universeD'$ and get relabelled with the belief \belief that justifies the action of \ego as a rational choice.
	We sketch the major steps of the construction as (i)-(iv) below:
	(i) We determine the current-state choices $\Act(\belief)$ for all beliefs in $\Beliefs$ by Prop.~\ref{prop:actBel}.
	(ii) We build the modified Kripke structure $\universeD'$: 
	Therefore we copy the state set \States of \universeD to become the state set $\States'$ of $\universeD'$.
	We then iterate over all states $\state\in\States$ of $\universeD$.
	If there is a transition from $\state$ via action $\act=(\act_1,\act_2)$ to $\state_2$ but no knowledge-consistent belief justifies \ego's action $\act_1$, \ie $\emptyset=\Beliefs_{\act,\state}:=\{\belief\in\Beliefs\mid \act_1\in{\Act(\belief)}$ and $\belief\models\LabelK(\state)\}$, we add a transition from \state to state \sundef and label this transition with $\act=(\perp,\act_2)$ to express that \ego will not choose this action, since it is no rational choice. %
	If $\Beliefs_{\act,\state}\not=\emptyset$, we iterate over all beliefs $\belief\in\Beliefs_{\act,\state}$ and introduce a transition from $\state'$ via $(\belief,\act_2)$ to $\state'_2$ in $\universeD'$, \ie we replace $\act_1$ by $\belief$.
	
	(iii) In order to judge how well the doxastic strategy \stratb has to perform for an autonomous-decisive system, we determine the maximal $\mn$ up to which $\goalList$ can be achieved by any \Props-truth-observing strategy in \universeD by iteratively applying strategy synthesis for LTL properties \cite{LTLSynth} starting from the maximum priority goals. 
	(iv) We then synthesize an \Obs-truth-observing strategy $\strats$ on {$\universeD'$}~\cite{LTLSynth} for the goal list $\goalList$ and priority $\mn$.
	In case $\strats$ achieves $\mn$, we define \LabelB by $\LabelB(\history):=\strats(\history)|_{\Beliefs}$, \ie the ${\LabelB}(\history)$ chooses the belief that labels the chosen transition.
	\stratbb may choose any action that is justified by $\LabelB(\history)$.
	Then $\sys=(\epi,\LabelB,\stratbb)$ is an optimal autonomous-decisive system. 
	If $\strats$ cannot achieve $\mn$, the truth-observing strategies on \universeD perform better, so no knowledge-consistent belief-formation for an autonomous optimal strategy exist for $\epi^{-}$.
\end{proofsketch}

To summarize, according to Theorem \autoref{th:auto} when designing an autonomous system \HAS, we can specify
\begin{itemize}
	\item the application domain via $\universeD$, 
	\item the list of goals \Goals, 
	\item the believed knowledge that the system \HAS will have, 
	\item what observations \HAS  can make and 
	\item how its internal representation the world is, \ie the possible worlds,
\end{itemize}
	and then we can determine whether it is at all possible to form beliefs such that the system \HAS is able to autonomously-decide and succeed as if it knew the ground truth. 
	Moreover, we can synthesize an appropriate belief labelling, so that the corresponding autonomous strategy is optimal for its goals. 

	Since we consider a quite liberal notion of autonomous system, it means that if the above check fails, it if often not possible to build an autonomous system with the given input and resources.  

Given we provide our system under construction full observability and let its beliefs reflect \universeD precisely and do not provide false believe knowledge, then an autonomous system \sys is guaranteed to exist.

The following example illustrates that the beliefs of an autonomous-decisive system \sys can be rather loosely linked to reality, observations are not (directly) represented in the beliefs and the possible worlds differ substantially from the ground-truth. 
Nevertheless, \sys can be successful. 

\begin{mexample}{Freedom of beliefs}
	In \autoref{fig:good} we sketch a belief formation where \Ego believes all the time, that \other has the wrong colour.
	The possible worlds do not reflect the ground truth world, \universeD, well, \eg in the possible worlds of $B_{0,1}$ and $B_{0,2}$ \other is red and the dominant strategy is not to turn, while in \universeD the dominant strategy is to do the turn, when \other is red. 
	Furthermore, the belief formation does rather abruptly update its beliefs (especially from $B_{0,2}$ to $B_{1,2}$).
	\begin{figure}[htbp]
		\centering
		\includegraphics[width=0.8\textwidth]{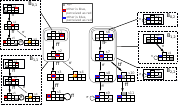}
		\caption{\Ego's sensor permanently switches colours. \Ego builds wrong and coarse beliefs that still enable it to act successfully.}\label{fig:good}
	\end{figure}
	Although the formed beliefs may seem degenerate and inappropriately capturing the reality, the belief formation allows \ego to behave as good as when it would know the ground truth. 
	This freedom of belief formation allows efficient and compact encodings of the perceptions comprised to the relevant aspects. 
	A system \sys that is optimal according to Def.~\ref{def:optimal} will have a belief formation that captures the application domain \universeD as closely as necessary to satisfy the system's goals.   

	If \universeD has to be captured even closer than necessary for \sys's goals, this can be enforced by means of the knowledge base.
\end{mexample}

\begin{mexample}{Autonomous, Non-Autonomous,  Automatic}\label{ex:autonAutom2}
	Let us consider an example of an autonomous \ego in the  setting of \autoref{fig:bellab}, where the sensor only initially switches colours, and consider the possible-worlds strategy \stratw of \autoref{ex:pws} (turn, if \other is hasty, and drive on, if \other is slow). 
	When \ego initially evaluates its situation in $s_1$ of \universeD, it believes that the situation is as described by {$B_{0,1}$}, i.e. \other is a hasty, red car.
	\Ego can decide to follow \stratw in {$B_{0,1}$}, as it seems a good choice -- \stratw is dominant and current-state decisive in {$B_{0,1}$}. 
	According to its extrapolation, it would move one step forward, and then it would successfully take the turn.
	After actually moving forward, \ego evaluates the situation in $s_2$.
	In $s_2$ \ego believes in {$B_{1,1}$} reflecting that \ego now truthfully perceives \other's colour as blue (cf. \autoref{ex:knowledgeBelief}).  
	Again, \stratw is a dominant current-state decisive possible-worlds strategy and determines \textsf{f} as the next move.
	Along this line, it is easy to see that \ego can implement a doxastic strategy \stratb that chooses the action that \stratw determines for the respective $\LabelB_{\tsf{auton}}(h)$.

	For an example for where no autonomous \ego exists but an automatic \ego can be built, we modify our running example slightly.
	Let us assume that \ego is unsure of its own initial position, thinking that it may initially be at \Small{$x=1$} or \Small{$x=2$}, as sketched in \autoref{fig:bellab3}. 
	Let a belief formation $\LabelB_{\tsf{autom}}$ be given that evolves the initial beliefs {$B_{0,3}$} and {$B_{0,4}$} analogously to \autoref{ex:knowledgeBelief}, that is, \ego perceives the correct colour after moving one step forward.
	The possible-worlds strategy \stratw is still a dominant strategy but not current-state decisive, since for example in {$B_{0,3}$} \ego would do the turn at $s1$ due to reality $r_2$, and it would also drive straight on due to $r_1$. Hence, there is no dominant possible-worlds strategy in {$B_{0,3}$} that is able to determine one action. \Ego cannot decide autonomously.
	Nevertheless, we can specify a dominant doxastic strategy \stratb for this case, but its actions are not chosen based on the belief content: \Ego chooses to turn after one step when its initial belief was {$B_{0,4}$} (\Small{$B_{0,4},B_{1,4}\mapsto t$}), otherwise it drives straight on.
	This strategy is dominant and could be used to build an \emph{automatic} system, where \ego just plays out \stratb. Such a strategy might be useful when an engineer knows that \ego will start from \Small{$x=1$} but did not equip \ego with this information.
\end{mexample}

A system that is not autonomous-decisive cannot rationally determine by itself which action is currently appropriate. A goal for the  design of a system \HAS is hence to ensure that a system is autonomous-decisive. 
\subsection{The Notion of Autonomous System}\label{sec:autoSys}
Above we have introduced our notion of \emph{autonomous-decisive system} as a system that takes rational decisions based on its current believes.
In Def.~\ref{def:optimal} we defined an \emph{optimal autonomous-decisive system}, as an autonomous-decisive system that performs at least as well as if it knew the ground-truth.
We do not intend these terms to characterize general autonomous systems nor do we aim  to capture aspects of free-will, independence or the ability of reflection; rather, we focus on the decision making of autonomous systems.
In the following we briefly discuss the notion of autonomous system and then relate our notions to it.

\paragraph{Autonomous Systems} The literal meaning of \emph{autonomy} is derived from \emph{auto = self} and \emph{nomos = law}.
Autonomy thus means self-governance \cite{merriamAutonomous} and the concept of autonomy can be found in different kinds of sciences~\cite{LitAutom2016}. 
For systems engineering the word autonomy describes the ability of a system to make its own decisions about its actions without the need for the
involvement of an outside supervisor~\cite{AutomStage}.

Although the terms \emph{automation} and \emph{autonomy} are sometimes used interchangeably~\cite{LitAutom2016},  
a significant difference
between the term autonomous and automatic is that an automatic system will do exactly as programmed while an autonomous system can make choices \cite{huAuto}. 
  
Several level of automation (LoA) have been suggested and discussed in literature.
Many of these see autonomy as the ultimate level of automation.~\cite{LitAutom2016} 
For instance, Parasuraman, Sherdian and Wickens list  in \cite{interactionLevel2000} ten levels of automation. 
Their levels target four broad classes of functions: information acquisition, information analysis, decision \& action selection and action implementation.
At the lowest level, humans must make all decisions and control all actions; at higher levels of automation, the automatic system increasingly takes over while humans receive less and less information about its operations. At level 10 the system decides everything and acts autonomously.\footnote{
In contrast, SAE J3016 defines a taxonomy for six levels of driving automation the SAE Levels of Driving Automation$^{\textsf{TM}}$ avoiding the term autonomy. 
They range from Level 0 (no driving automation) to Level 5 (full driving automation) in the context of motor vehicles and their operation on roadways \cite{SAE}.}

According to J. Sifakis, \cite{AutoSysSifakis}, the main characteristic of autonomous systems is their
ability to handle knowledge and adaptively respond to environment changes.
Autonomous systems have to operate for extended periods of time under
significant uncertainties in the environment and they have to compensate a certain amount of system failures, both without external intervention \cite{TwIntAutCS1989}. 
Many agree with ~\cite{AutoSysSifakis} that autonomy combines perception, reflection, goal management, planning and self-adaptation~\cite{AutoSysSifakis}.
Often autonomous systems are discussed with a focus on artificial intelligence and learning \cite{industAuto,AutoSysSifakis}.

\paragraph{Autonomous-decisive Systems} In what follows, we argue that our notion of optimal autonomous-decisive system fits well with the notion of autonomous system as outlined above.

The key asset of our notion is formalizing an \enquote{epistemic goal-directedness} for autonomous systems.
Our notion is based  on the fact that a system perceives its environment and maintains a varying knowledge base.
We introduce the explicit requirement that the devised plans have to be rational with respect to its internal world view.  
Thereby we can also distinguish autonomous systems from automatic systems in a way that is compatible with \eg \cite{huAuto}.

What about reflection, goal management, planning and self-adaptation? 
Our work is primarily concerned with decision making. We see you framework as a first step towards formalizing and studying a couple of interesting properties of autonomous systems, as we sketch below.
 
 The formal framework does not constrain what kind of information the \HAS's beliefs encode nor 
what kind of actions an autonomous systems can perform or how it perceives feedback regarding its actions' effect. 
We hence believe that the framework allows to study systems that have explorational awareness, \ie that explore the environment gathering information as part of their strategy.
For instance a robot can explore unknown paths of maze by keeping track where it has been.

Similarly, we believe that reflection can be treated within this framework. 
To this end, the \HAS's believes have to encode believes on believes -- not only represent the believed factual world.
We imagine that finite believe hierarchies could be encoded similar to \cite{perea2012}.
While the treatment of explorative system seems within the framework seems to be straight-forward, modeling believe-hierarchies is considered as future work. 

The framework as such that does not have a notion of goals or is concerned with goal management. 
We believe that the framework could be extended by a notion of subgoals, in such a way that it is possible to analyze whether there is a strategy for subgoal selection.
Conceptually, goals of autonomous system seem to be a mean for breaking down a complex global goal to more easily treatable goals.
So instead of subgoal selection, we can examine whether there is a global strategy that depends only on a certain limited simulation and planning horizon.

Finally, we want to remark that we can consider the design time universe as a training set where certain known aspects of the world are captured. A deployed \HAS then has to be equipped with a belief formation that is able to deal with unexpected events. 
Studying formal robust properties of the belief formation seems an important and interesting endeavour, \eg \enquote{Given a belief formation, how much can the real world deviate from the design-time world?} or \enquote{How much timing tolerance does a certain belief formation have?}.

\section{Relevance}\label{sec:relevance}
We consider the combination (\LabelK,\Obs,\Beliefs) of labeled knowledge, observations and possible beliefs as important set screws for an engineer to develop an optimal autonomous-decisive system.
 \Small{{{\RE}\,{\LabelK}}}, he can equip the system \HAS with prior knowledge and implement mechanisms to update \HAS's knowledge base during the mission, 
 \Small{{{\re}\,{\Obs}}}, he can provide more sensing capabilities and, 
 \Small{{{\re}\,{\Beliefs}}}, he can increase the resources for the internal representation of the world model.
In the following we denote (\LabelK,\Obs,\Beliefs) also as \kob.

To support an engineer, we characterise whether a tuple \kob is sufficient for a given setting.
The basic idea is: If a system is an \emph{optimal} autonomous system, then its formed beliefs conserve the relevant aspects of \universeD. 
Hence the \kob is sufficient if a relevance conserving belief formation exists. 
To answer whether \kob are relevant, we test whether it is possible to build an optimal autonomous system with less knowledge, observations or beliefs. 

For the following we consider a doxastic model $\epi$ to be given with $(\universeD,\goalList,\LabelK,\Obs,$ $\Beliefs,\LabelB)$ with a knowledge-consistent belief formation \LabelB and a system $\sys=(\epi,\stratb)$ with a doxastic strategy.\\

\subsection{Conservation of the Relevant}
We first define when the relevant is conserved. 
Therefore we compare \ego's (doxastic and autonomous-decisive) performance with the performance that \ego could have when it would access the ground-truth, \universeD.

We first develop a notion of relevance conservation for doxastic systems in order to highlight that the requirements for autonomous systems are more demanding. \\

We say that the belief formation \LabelB conserves the relevant of \universeD, 
if \epi can perform based on its beliefs as successful as it could when directly and truthfully observing the ground-truth \universeD. %
\begin{mdefinition}{Relevance Conservation for Doxastic Systems}\label{def:wrelcon}
Let $\Obs_D\subseteq\Props$ be a set of propositions.
	The belief formation $\LabelB$ of a doxastic model $\epi=(\universeD,\goalList,\LabelK,\Obs,\Beliefs,\LabelB)$ \emph{conserves the relevant of a $\Obs$-observable \universeD}, if there exists a doxastic strategy \stratb  for \epi that is dominant \wrt all $\Props$-observing strategies $\strats$.
\end{mdefinition}
When $\LabelB$ is conserving the relevant of  completely observable design-time model \universeD. 
then \ego could --by implementing \stratb of Def.~\ref{def:wrelcon}-- perform as well as possible when the ground-truth \universeD would be completely observable. 
Def.~\ref{def:wrelcon} captures this aspect by comparing the performance of \ego that is observing \Obs with the performance on the ground-truth \universeD that is observable via $\Props$.  

But what does it mean that a belief formation $\LabelB$ conserves the relevant? Intuitively, it means that $\LabelB$ preserves \universeD in sufficient detail to map the history of beliefs \enquote{somehow} to the best action.
The choice of action does not have to be plausible \wrt the content of a system's beliefs though. 
It is up to the engineer to choose which strategy $\stratb$ will be implemented by the system.\\

The choice of action must be plausible \wrt the belief content though, when it comes to autonomously-decisive systems.
An autonomous-decisive system chooses at all times actions \act that are justified in the respective current belief \belief, \ie $\act\in\Small{\Act(\belief)}$ (cf. Def.~\ref{def:choiceb}).
We hence say that the belief formation conserves the relevant for autonomous-decisiveness, if at all times the \enquote{best actions} \wrt the belief content are chosen. 
\begin{mdefinition}{Relevance Conservation for Autonomous-Decisiveness}\label{def:relcon}
	Let $\Stratsbb$ be the set of autonomous strategies that exist for $\epi=(\universeD,\goalList,\LabelK,\Obs,\Beliefs,\LabelB)$.

	The belief formation $\LabelB$ of \epi \emph{conserves the relevant of a $\Obs$-observable design-time world \universeD for autonomous-decisiveness}, 
	if all $\stratbb\in\Stratsbb$ are dominant \wrt $\Props$-observing strategies $\strats$ on \universeD.
\end{mdefinition}
Note, that we assume (\cf Ass. \ref{ass:csdec}, p.~\pageref{ass:csdec}) that the belief formation \LabelB forms only beliefs \belief in which a dominant current-state decisive strategy exists. Hence  $\Stratsbb$ contains at least one strategy.

 If \LabelB conserves the relevant for autonomous-decisiveness as defined in Def.~\ref{def:relcon}, then any of the autonomous strategies \stratbb of \epi observing \universeD via \Obs will perform  as successful as possible when directly accessing \universeD via $\Obs'$. 
 It may seem surprising, that Def.~\ref{def:relcon} refers to \emph{all} autonomous strategies $\stratbb\in\Stratsbb$. The reason is, that the final decision on the chosen action lies with the autonomous system. 

\begin{mexample}{Conservation of the Relevant}\label{ex:conserve}
	As an example of a belief formation that conserves the relevant for autonomous-decisiveness, we refer the reader back to  \autoref{ex:autonAutom2} on page \pageref{ex:autonAutom2}. 
There we sketched a setting where the sensors are initially broken but when the decision has to taken the sensors provide the relevant information. 
The resulting belief formation $\LabelB_{\textsf{auton}}$ allows a system to perform as well as when knowing the ground-truth, \ie not having initially disturbed sensor readings. 
	In  \autoref{ex:autonAutom} on page \pageref{ex:autonAutom} we saw an example of a belief formation that conserves the relevant for doxastic systems but not relevance for autonomous-decisiveness. 
	In the example, \ego's sensors are permanently switching colours and \ego has a knowledge base that forces it to believe that a read car is fast and a slow car is blue. 
	Consequently, \ego cannot autonomously determine what is best to do in \universeD. 
	But the belief-formation is such that an engineer can choose a strategy for \ego that deals with the color readings appropriately, \ie \enquote{switch them back}.
\end{mexample}

Conserving the relevant for autonomous-decisiveness is stronger than conserving the relevant for doxastic systems: 
\begin{proposition}[Relevance Conservation]\label{th:refine}
	\enspace\\[-2mm]\enspace
	\begin{enumerate}
		\item\label{it:autDox} If \LabelB conserves the relevant for autonomous-decisiveness, then \LabelB conserves the relevant for doxastic systems.
	
		\item\label{it:doxAut} If \LabelB conserves the relevant for doxastic systems, then \LabelB does not necessarily conserve the relevant for autonomous-decisiveness.
	\end{enumerate}
	\end{proposition} 
\begin{proofsketch}{Prop. \ref{th:refine}}
	Prop.~\ref{th:refine}(\ref{it:autDox}) follows directly from Def.~\ref{def:wrelcon} and Def.~\ref{def:relcon}. Prop.~\ref{th:refine}(\ref{it:doxAut}) follows from the example \ref{ex:conserve}.
\end{proofsketch}

The next proposition is concerned with systems, where the belief formation is captured via a set of rules. 
Such autonomous systems still play an important role especially in safety critical applications, although artificial intelligence systems, that intransparently build their beliefs, gain more and more importance.

Since the resources of a system \HAS are limited, we consider belief formation functions that can be represented by a finite number of regular expressions. 
\begin{mdefinition}{Regular Belief Formation}
We say \LabelB is regular, if $\LabelB$ can be defined via a finite number $n$ of regular expressions $\regEx_i$, i.e., 
for all observable histories $\history\in\histories_{\Obs}$ of \universeD it holds, that there is an $i, 1\leq i\leq n$ such that $\LabelB(\history)=\belief_i$ iff $\history\models\regEx_i$.
\end{mdefinition}

Given a doxastic model with a regular belief formation \LabelB, we can decide whether \LabelB conserves the relevant for autonomous-decisiveness:
\begin{proposition}[Conservation of Relevance]\label{th:check}
	Given a regular belief formation \LabelB, we can decide whether \LabelB conserves the relevant for autonomous-decisiveness.
\end{proposition}
\begin{mproof}{Prop. \ref{th:check}}
	We first determine the maximal priority $\mn$ up to which the goal list $\goalList$ can be achieved on \universeD by applying iteratively strategy synthesis for LTL properties~\cite{LTLSynth} starting with the maximum goal list and then iteratively decreasing $\mn$. 	
	\LabelB of $\epi$ conserves the relevant, if all autonomous strategies $\stratbb\in\Stratsbb$ achieve at least $\mn$ (\cf~Def.~\ref{def:relcon}).
	We then construct an automaton $\aut_{\Act()\times\universeD}$, in which the environment is unconstrained and \ego chooses its actions from $\Act(\LabelB(h))$ after observing history $h$.
	It holds that iff $\aut_{\Act()\times\universeD}$ satisfies $\goalList$ up to $\mn$, then  \LabelB conserves the relevant for autonomous systems.

	Construction of $\aut_{\Act()\times\universeD}$: For each belief $\belief\in\Beliefs$, we can determine the current-state choices $\bAct(\belief)$ (Prop.~\ref{prop:actBel}). 
	Thus, the belief formation \LabelB can be considered as an \Obs-observing strategy assigning $\Act(\belief)$ to an observed history \history with $\belief=\LabelB(\history)$: 
	Since \LabelB is regular, we can build a mealy automaton $\aut_{\Act()}$ that determines $\Act(\LabelB(\history))$ for an observed history \history.
	When $\aut_{\Act()}$ transitions to an accepting state because of \history, this transition gets labelled with the current-state choices $\Act(\LabelB(\history))$.
	We derive a composed automaton ${\aut_{\Act()\times\universeD}}$ by parallel composition of the design-time world \universeD and $\aut_{\Act()}$.  
	In $\aut_{\Act()\times\universeD}$, \ego can take an action \act only if  $\aut_{\Act()}$ allows this, i.e. it is a current-state choice for the observed history. 
	If \ego may not take $\act_1$ in state \state, the combined action $\act=(\act_1,\act_2)$ for all $\act_2\in \Act_{\env}$, leads to the state \sundef. 
\end{mproof}

In the next section we will characterise what knowledge, observations and beliefs are relevant.
Thereby we turn to questions like \enquote{Can we do with less observations?}, \enquote{Can we do with less detailed beliefs?} or \enquote{Can we compensate missing observations by adding knowledge?}.
%
\subsection{Relevance of \kob}
%
Our notion of \emph{relevance conservation} characterises combinations of \kob that allow  a system to form beliefs that are \emph{sufficiently precise} for the system to be optimal. 
In this section we define that \kob is \emph{relevant}, if it conserves the relevant (\ie is sufficient), and in additional also \emph{\enquote{minimal}}. 

The three dimensions of \kob are of course interrelated. Intuitively, knowledge (\LabelK) about the world can replace observations that a system \HAS needs otherwise. 
Having more resources for the representation of the inner world model (\Beliefs) allows a system \HAS to store more of the made observations and allows it  to make finer predictions.
More observations (\Obs) vice versa allow \HAS to forget more  and thus have a simpler model of the history and future or to have less knowledge.
We hence expect that often several incomparable minima can be determined. 

To define a \emph{minimal \kob}, we first define partial order relations on  the set of knowledge labeling functions, the set of observations and the set of possible beliefs. 
We then infer a partial order to order tuples \kob.

We chose the partial orders to reflect the decisions an engineer has to make during the design:
\begin{enumerate}[label=\small(PO\arabic*)\normalsize,itemindent=4mm]
	\item\label{i:obs}  $\Obs\leq\Obs'$ $:\Leftrightarrow$ $\Obs\subseteq\Obs'$\\
For this paper we assume that a greater set of observations means that more sensors are necessary. We are hence interested in determining the minimal set of required observations. 
	\item\label{i:bel}  $\Beliefs\leq\Beliefs'$ $:\Leftrightarrow$ $\Beliefs\subseteq\Beliefs'$\\
		For the design of a \HAS the size of the set of possible beliefs \Beliefs corresponds to the resources that are necessary to encode the beliefs.
	\item\label{i:know} $\LabelK\leq\LabelK'$ $:\Leftrightarrow$ $\forall \state\in\States: \LabelK(\state) \leq \LabelK'(\state)$ $:\Leftrightarrow$ $\forall \state\in\States: [\LabelK'(\state)]\subseteq [\LabelK(\state)]$,\\
		where [{\priorK}] denotes the set of traces on all possible worlds, $\Worlds$, that satisfy the believed knowledge $\priorK$. 
		As we deal with knowledge-consistent belief formations here, $\priorK\leq\priorK'$ means that \priorK' constrains the beliefs that can be formed less. 
		In other words, the system \HAS knows less since it has more uncertainty.
		
		$\LabelK\leq\LabelK'$ means that $\LabelK'$ declares more knowledge at least at one state of \universeD and it declares not less knowledge than \LabelK in all other states. 
		An engineer can provide prior knowledge, \eg she can hard-code the believed knowledge into \HAS, and she can implement the knowledge labelling, i.e. ensure that mechanisms are in place that will update the knowledge base during \HAS's missions.
	\item\label{i:kob} $\kob \leq \kobp$ $:\Leftrightarrow$ \ref{i:obs}-\ref{i:know} hold.

\end{enumerate}
By  \ref{i:kob} we now  define the notion of \emph{weak relevance}. 
A tuple \kob is weak relevant, if we cannot find a strictly smaller tuple \kobp.
We call this \emph{weak}, since there can be other tuples \kob that are incomparable with \kob.
Hence the question \enquote{Is \kob relevant} does not have a definite answer.
Nevertheless the notion of weak relevance allows to answer, whether a system \HAS can do with more observations in exchange for less knowledge or fewer possible beliefs or whether more knowledge allows \HAS to have fewer possible beliefs or less observations and so on. 

\begin{definition}[Weak Relevance]\label{def:wr}
	Let a design-time world $\universeD$ and a prioritised list of goals $\goalList$ be given.

	\kob is weakly relevant for $(\universeD,\goalList)$, if 
	\begin{enumerate}
		\item\label{def:cons} there is a belief formation \LabelB of $\epi:=(\universeD,\goalList,\LabelK,\Obs,\Beliefs,\LabelB)$ that conserves the relevant for autonomous systems and 
		\item\label{def:min} for all $\kobp\not=\kob$ with 
			$\LabelK'\leq\LabelK$, $\Obs'\leq\Obs$ and $\Beliefs'\leq\Beliefs$
			$\kobp$  there is no knowledge-consistent belief formation \LabelB' of $\epi':=(\universeD,\goalList,\LabelK',\Obs',\Beliefs',\LabelB')$ that conserves the relevant for autonomous systems.
	\end{enumerate}
	\LabelK is weakly relevant if there are \Obs and \Beliefs, such that \kob is weakly relevant. 
	Analogously we define that \Obs (\Beliefs) is weakly relevant if there are \LabelB, \LabelK and \Beliefs (\Obs). 
\end{definition}
To illustrate the notion, we consider an example. 
\begin{mexample}{Weak Relevance}
Let us assume \ego observes its position $\pos$, a time stamp $t$ and its speed $v$. Its goal is to determine its past average acceleration $\acc$.%
\footnote{We assume finite domains and hence finite encodings of numerical values. The computations will be rounded appropriately. \Ego's actions are computation steps.} 
Moreover, let us assume that the perception of position is flawed when it is raining while the speed is still correctly measured.  
Then only $\{v,t\}$ is weakly relevant, that is, they suffice to determine the average acceleration. 
	Neither the set $\{\pos,v,t\}$ is weakly relevant nor the set $\{\pos,t\}$. The further is not minimal, the latter does not conserve the relevant, since $\acc$ cannot be determined while it is raining.

	Given \ego has the knowledge \enquote{it will not rain} both $\{\pos,t\}$ and $\{v,t\}$ are weakly relevant. 
\end{mexample} 

Let us now turn to questions  like \enquote{Is \Obs relevant, given \LabelK and \Beliefs?}, i.e. we assume tow component of the triple are known.
The question of relevance ten might have a definite answer, but not necessarily. We hence consider it an interesting notion.
In Def.~\ref{def:relevance} we define \LabelK (\Obs, \Beliefs) to be relevant, if there is no alternative minimal choice, i.e., the system \HAS has to have \LabelK (\Obs, \Beliefs) in order to be able to perform autonomously optimal.
\begin{mdefinition}{Relevance}\label{def:relevance}
	\Obs is relevant for (\universeD,\goalList) with (\LabelK,\Beliefs) iff
	\begin{enumerate}
		\item \Obs is weakly relevant and 
		\item there is no other \Obs' that is weakly relevant.
	\end{enumerate}
	Likewise we define \LabelK and  \Beliefs are relevant for (\universeD,\goalList) with (\Obs,\Beliefs) and respectively (\LabelK,\Obs). 
\end{mdefinition}
\begin{mtheorem}[Relevance]\label{th:relevant}
	Given a doxastic model   $\epi=(\universeD,\goalList,\priorK,\Obs,\LabelB)$ of \HAS within its environment, we can decide whether \ko is (weakly) relevant for \LabelB in \epi. 
\end{mtheorem}	
\begin{mproof}{\autoref{th:relevant}}
	To show \autoref{def:cons} of Def.~\ref{def:wr} we check whether there is a \Obs-observing strategy in \universeD. 
	To check \autoref{def:min} of Def.~\ref{def:wr} we build the \enquote{lesser} pairs $(\priorK',\Obs')$, i.e. $\priorK'\leq\priorK$, $\Obs'\leq\Obs$ and $(\priorK,\Obs)\not=(\priorK',\Obs')$, and check whether there is a belief labelling \LabelB' that conserves the relevant again by checking whether there a $\Obs'$-observing strategy in \universeD.
\end{mproof}

\bibliographystyle{plain}
\bibliography{refs}
\end{document}